%% file: main.tex
\documentclass[11p]{article} 
\def\arXiv{1}

\newcommand{\notarxiv}[1]{foo}
\newcommand{\arxiv}[1]{ba}
\ifdefined \arXiv
	\renewcommand{\arxiv}[1]{#1}%
	\renewcommand{\notarxiv}[1]{\ignorespaces}%
\else%
	\renewcommand{\arxiv}[1]{\ignorespaces}%
	\renewcommand{\notarxiv}[1]{#1}%
\fi%

\arxiv{
	\usepackage[top=1in, right=1in, left=1in, bottom=1in]{geometry}
	\usepackage{url}
	\usepackage[hidelinks]{hyperref}
	\hypersetup{
		colorlinks=true,
		linkcolor=blue!70!black,
		citecolor=blue!70!black,
		urlcolor=blue!70!black}
	\usepackage[numbers, square]{natbib}
}

\notarxiv{
	\usepackage[subtle, all=normal, mathspacing=normal, floats=tight, 
	sections=tight]{savetrees}
	\usepackage[hidelinks=true]{hyperref}
}

\usepackage[utf8]{inputenc} 
\usepackage[T1]{fontenc}    
\usepackage{hyperref}       
\usepackage{authblk}
\hypersetup{
    colorlinks=true,
    citecolor=blue,
    linkcolor=blue,
    filecolor=magenta,     
    urlcolor=blue,
}
\usepackage{url}            
\usepackage{booktabs}       
\usepackage{amsfonts}       
\usepackage{nicefrac}       
\usepackage{microtype}      
\usepackage{xcolor}         

\input{math_commands.tex}

\input{macros.tex}

\input{paper_macros}
\usepackage{cleveref}
\usepackage{algorithm}
\usepackage{algpseudocode}
\usepackage{subcaption}
\usepackage{wrapfig}
\usepackage{array}
\usepackage{makecell} 
\usepackage{wrapfig}
\usepackage{thmtools}
\usepackage{thm-restate}

\usepackage{tablefootnote}
\usepackage{centernot}
\usepackage{tikz}
\usepackage{accents}
\allowdisplaybreaks

\usepackage{multirow}
\usepackage{diagbox}

\usepackage{color}
\usepackage{soul}
\usepackage{pgfplots}

\title{Time After Time: Deep-Q Effect Estimation for Interventions on When and What to do\thanks{accepted for presentation at the International Conference on Learning Representations (ICLR) 2025}}

\author[1]{Yoav Wald\thanks{Correspondence to \texttt{yoav.wald@nyu.edu}}}
\affil[1]{New York University}

\author[1]{Mark Goldstein}
\author[2]{Yonathan Efroni}
\affil[2]{Meta}
\author[3]{Wouter A.C. van Amsterdam}
\affil[3]{University Medical Center Utrecht}
\author[1]{Rajesh Ranganath}

\date{}

\begin{document}

\maketitle

\begin{abstract}

  
Problems in fields such as healthcare, robotics, and finance requires reasoning about the value both of what decision or action to take and when to take it. The prevailing hope is that artificial intelligence will support such decisions by estimating the causal effect of policies such as how to treat patients or how to allocate resources over time. However, existing methods for estimating the effect of a policy struggle with \emph{irregular time}.  They either discretize time, or disregard the effect of timing policies.
We present a new deep-Q algorithm that estimates the effect of both when and what to do called Earliest Disagreement Q-Evaluation (EDQ). 
EDQ makes use of recursion for the Q-function that is compatible with flexible sequence models, such as transformers. EDQ provides accurate estimates under 
standard assumptions.
We validate the approach through experiments on survival time and tumor growth tasks.
\end{abstract}

\input{sections/1_introduction}
\input{sections/2_setting}
\input{sections/4_estimation}

\input{sections/3_related_work}

\input{sections/3B_related_work_technical}
\input{sections/5_experiments}

\bibliography{main}
\bibliographystyle{apalike}

\newpage
\appendix
\section{Additional Comments on Experimental Aspects} \label{app:experiments_comments}
Here we slightly expand on the comment about computational complexity in the main text, and give more details about the cancer simulation we use from \citet{geng2017prediction, Bica2020Estimating, seedat2022continuous,vanderschueren2023accounting}.

\emph{A comment on computational complexity:} As commented in the main text, the per-iteration runtime of \ours{} is similar to that of FQE, which is a common tool in large-scale offline RL problems; for example, \citet{paine2020hyperparameter, voloshin2021empirical} use it in benchmarks and evaluations. The difference in computation times between \ours{} and FQE is due to sampling from the target policy, or more accurately $\tilde{P}_t^a$, in order to draw the treatments used in the $Q$-update, i.e., $\delta$ and $\tgH_{t+\delta}$ in \cref{alg:FQE_earliest disagreement}. In most applications, the added complexity due to this difference is small relative to the cost of evaluating the $Q$-function and its gradients. In turn, the cost of function evaluation is the same for FQE and \ours{}. The computational complexity of sampling from $\tilde{P}_t^a$ depends on how it is represented and implemented. For instance, we may specify policies by allowing evaluations of $\lambda^a(u \vert \gH_u)$, and sample using the thinning algorithm \citep{lewis1979simulation, ogata1981lewis}; with neural networks that allow sampling the time-to-next-event (e.g., see \citep{nagpal2021deep, mcdermott2023event} for examples of event time prediction); or with closed-form decision rules. For example, in the time-to-failure simulation, to determine treatment times we sample exponential variables every times the vital feature crosses a certain threshold.

\emph{Cancer simulator:} The tumor growth simulation we use is adapted from \citet{Bica2020Estimating, seedat2022continuous, melnychuk2022causal} and is based on the work of \citet{geng2017prediction}. Tumor volumes $V(t)$ are simulated as finite differences from the following differential equation,
\begin{align*}
\frac{d V(t)}{d t} = \left( \underbrace{\rho \log \left( \frac{K}{V(t)} \right)}_{\text{Tumor growth}} - \underbrace{\beta_c C(t)}_{\text{Chemotherapy}} - \underbrace{\left(\alpha_r d(t) + \beta_r d(t)^2 \right)}_{\text{Radiotherapy}} + \underbrace{e_t}_{\text{Noise}}\right)V(t).
\end{align*}
Here $C(t)$ is the chemotherapy concentration, $d(t)$ represents the level of radiothearpy. $\rho, K, \beta_c, \alpha_r, \beta_r$ are effect parameters drawn for each patient from a prior distribution described in \citet{geng2017prediction}, and $e_t\sim \mathcal{N}(0, 0.0001)$ is a noise term. To create irregularly sampled observations of the tumor volume, at each time step we draw a value from a Bernoulli distribution to decide whether the trajectory contains the tumor volume at this time step or not. The success probability is a function of the average tumor volume over the most recent $15$ volumes (both observed and unobserved). If we denote a missing value by $\emptyset$ and the observation at timestep $t$ by $X_t$ (which equals $\emptyset$ if there is no sample at this timestep and $V(t)$ otherwise), then sampling times are drawn according to the following probabilities:
\begin{align*}
    p(V_t \neq \emptyset \vert \gH_t) = \sigma\left(\frac{\bar{V}_{t-15:t}}{V_{\max}} - 1.5\right)
\end{align*}

The policies we use to decide on treatments draw binary decisions of whether or not to apply chemotherapy and radiotherapy at each timestep. Denoting these decisions by random variables $C_t$ and $d_t$, they are drawn according to $P(R_t = 1 \vert \gH_t) = \sigma\left( \gamma\cdot \left( v_{\text{last}} - \beta \right) + t - t_{\text{last}} \right)$, where $v_{\text{last}}$ is the last observed volume before time $t$ and $t_{\text{last}}$ is the last time that treatment was applied before $t$. The same probabilities are applied for $d_t$.
For more details on the specifics of the simulation, see the code implementation. 
\input{sections/A_proofs}

\section{Additional Discussion on Identifiability} \label{app:identifiability}
To make the discussion on identifiability from \cref{sec:identifiability} more concise, we refer here to terms and results from \citet{roysland2022graphical}, and explain how they apply to our setting. Then we prove \cref{corro:EDQ_estimate}
\subsection{Causal Identification Terminology}
\paragraph{Filtrations and restrictions.} In our notation, we condition on past events $\gH_t$, while in most formal treatments of stochastic processes conditioning is performed on a filtration $\gF_t$ of a $\sigma$-algbera $\gF$. $\gF$ is the full set of information generated by the processes $N=[N^a, N^x, N^y, N^u]$. The notation $\gF^{v}$ is used for the $\sigma$-algebra generated by the process $N^v$ alone, hence to a reduced set of information with regards to $\gF$. Note that here we use the index $v$ instead of $e$ which we used in the main paper. This is to avoid confusion with the notation for edges in a graph $G$. In the notation of \citet{roysland2022graphical}, there are also corresponding filtrations $\gF^v_t$ of $\gF^v$. Finally the restriction $P\vert_{\gF^v}$ is used to denote the restriction of probability measure $P$ to the sub $\sigma$-algebra $\gF^v$. Intuitively, $P\vert_{\gF^{x,a,y}}$ is the measure that ignores all the information generated by the unobserved processes $N^u$ by marginalizing them. Outside this section, \cref{app:identifiability}, the notation $P$ is used to refer directly to $P\vert_{\gF^{x,a,y}}$ as we work under the ignorability assumption, which means inference under the restricted probability distribution yields valid causal effects. In the rest of this section we explain why this validity holds, hence $P$ refers to the full process with the unobserved information included.

Following the above discussion on filtrations, expressions such as $\lambda^v(t \vert \gH_t)$ should be read as conditioning on a subset of the sample space where the events in the time interval $[0, t)$ coincide with $\gH_t$.
\paragraph{Interventions and causal validity.} Under the assumption on independent increments, item $2$ in Assumption~\ref{assump:ignor}, we have that densities for a trajectory $\gH$ take on the form:
\begin{align*}
    P_{\obs}(\gH) = \exp{\left( -\int_{0}^{T}\lambda_{\obs, \bullet}(t \vert \gH_{t}) \mathrm{d} t \right)}\prod_{z\in{\{a, x, y\}}, t_{z,i}\in{\gH^z}}{\lambda^z_{\obs}(t_{z,i} \vert \gH_{t_{z,i}})}.
\end{align*}
Interventions on $N^a$ then mean we replace the intensity $\lambda^a_{\obs}$ (which may depend on $\gH^u$) with an intensity $\lambda^a$ that only depends on $\gH^{x,a,y}$ (formally, this means it is $\gF^{x,a,y}$ predictable). The densities change accordingly

The question of causal validity is then whether calculating statistics under $P$, e.g. $E_P[Y \vert \gH^{a,x,y}_{t}]$, results in the same estimation as calculating them under $P\vert_{\gF^{a,x,y}}$. The difference being that $P\vert_{\gF^{a,x,y}}$ intervenes on $P_{\obs}\vert_{\gF^{a,x,y}}$ instead of on $P_{\obs}$.\footnote{\citet{roysland2022graphical} also mention the assumption that $P$ itself is causally valid, i.e. that calculating effects under the distribution with the unobserved variables indeed yields valid causal effects. Here we take this assumption as a given.} The condition of eliminability, along with overlap, ensures that this validity holds.

\paragraph{Eliminability, local independence, and the assumed graph.} 
We will local independence graph $G$ is a directed graph where each node
We now recall the definition of eliminability and the result we use from \citet{roysland2022graphical}.

$P\vert_{\gF^{x,y,a}}$ is the distribution that performs the intervention on $P_{\obs}$ and then restricts information to the observable information, which admits the causal effect we wish to estimate. We then denote $\tilde{P}$ as the distribution obtained by the same intervention on $P_{\obs}\vert_{\gF^{x,y,a}}$ (i.e. where we marginalize over $U$ before the intervention).
We rephrase the theorem of \citet{roysland2022graphical} to make it more compatible and specialized to our notation and case, it is strongly advised to review the results in their paper for a full formal treatment.
\begin{theorem}[special case of \citet{roysland2022graphical}]
Let $P(G)$ be a local independence model and the nodes be partitioned as in \cref{def:elim}. Consider $P_{\obs}\in{\mathcal{P}(G)}$ and a distribution $P$ obtained by an intervention on the process $N^a$ in $P_{\obs}$, replacing
its $\gF$-intensity $\lambda_{\obs}^a$ by a $\gF^{x,y}$-intensity $\lambda$. Further consider $\tilde{P}$ that is obtained by performing the intervention on $P_{\obs}\vert_{\gF^{x,a,y}}$.

If $G$ satisfies eliminability, then the intensities $\lambda$ of $P\vert_{F^{a,x,y}}$ coincide with the intensities of $\tilde{P}$.
\end{theorem}

To finish this overview, we explain in detail why the graph we assume in this work satisfies eliminability, and hence under the additional assumption of overlap, i.e. Assumption~\ref{assump:overlap}, we conclude the correctness of our estimation technique.
\begin{definition}
A trail from $v_0$ to $v_m$ in $G=(V, E)$ is a unique set vertices $\{v_0, \ldots , v_m\}\subseteq{V}$ 
and edges $\{e_1, \ldots, e_m\}\subseteq{E}$
such that either $e_j = v_{j-1} \rightarrow v_j$
or $e_j = v_{j-1} \leftarrow v_j$ for every $j = 1, \ldots, m$.

The trail is \emph{blocked} by $C \subseteq V$ if either (i) $V$ contains a
vertex on the trail that is not a collieder (i.e. there is no $j$ such that both $e_j = v_{j-1} \rightarrow v_j$ and $e_{j+1} = v_j \leftarrow v_j$), or (ii) $v_j$ is a collider for some $j\in{[m]}$, while $v_j\notin{C}$ and $v\notin{C}$ for any $v$ which is a descendant of $v$.

The trail is \emph{allowed} if $e_m = v_{m-1} \rightarrow v_m$. We say that $A\subseteq{V}$ is $\delta$ separated from $u$ by $C$ if for any $a\in{A}$, $\{u\} \cup C$ blocks all allowed trails from $a$ to $u$.
\end{definition}
\citet{didelez2008graphical} gives results that tie $\delta$-separation to \emph{local independence}, namely that under some regularity conditions if $A$ is $\delta$-separated from $v$ by $C$, then $v$ is locally independent of $A$ given $C \cup {v}$. Local independence in turn is a condition on the intensities of the process, namely that the $\gF^{\{v\}\cup{C}\cup{A}}$ intensity of $N^v$ is indistinguishable from its $\gF^{\{v\}\cup{C}}$-intensity. Intuitively, the $\gF^{\{v\}\cup{C}}$-intensity is the intensity that does not include information from the past of $N^A$. Formally, it may be obtained using the innovation theorem \citep[II.4.2]{andersen2012statistical}. Then to show that the graph we assumed in our derivation satisfies eliminability and conclude our claims, we will explain why the appropriate $\delta$-separation properties hold.
\begin{claim}
The graph $G$ in \cref{fig:local_graph} satisfies eliminability.
\end{claim}
\begin{proof}
Consider $U_1$, it is easy to verify that $N^a$ is locally independent of $U_1$ given $N^x, N^y$
(the second option from \cref{def:elim}). This is because $N^x, N^y$ blocks all directed paths between $U_1$ and $N^a$ on the graph, and in paths where one of these nodes is a collider, the other is not. Next we consider $U_2$, and claim that $(N^y, N^x)$ are locally independent of $U_2$ given $(N^x, N^y, N^a)$, which means the first bullet in \cref{def:elim} is satisfied. To see this local independence holds, consider any allowed path in $G$ from $U_2$ to $N^x$ or $N^y$. Since allowed trails must end with incoming edges to the last node, then either $N^a$ must be a non-collider on such an allowed trail (e.g. in the trail $U_2 \rightarrow N^a \rightarrow N^x$), or either $N^x$ or $N^y$ must be a non-collider (e.g. in the trail $U_2 \rightarrow N^a \leftarrow N^y\rightarrow N^x$). In both cases, these non-colliders are in the conditioning set, and thus $U_2$ is $\delta$-separated from $N^x, N^y$ by $(N^a)$. The required local-independence follows from this.\footnote{the condition also includes the independence of $U_{>2}$, but in our case $U_{>2} = \emptyset$.}
\end{proof}
\subsection{Proof of \cref{corro:EDQ_estimate}}
Equipped with the proper definitions of identifiability, we can now conclude our discussion on \ours{}. Let us recall \cref{corro:EDQ_estimate}.
\edqcorro*

Following our discussion on identifiability, we see that assumption~\ref{assump:ignor} guarantees that calculating conditional expectations w.r.t a distribution where we replace $\lambda^a_{\obs}$ by $\lambda^a$ in $P_{\obs}\vert_{\gF^{x,a,y}}$, yields a correct causal effect. For the rest of this section we will call the resulting distribution $P$. This is a slight abuse of notation, since $P$ was used to refer to the interventional distribution under on $P_{\obs}$, but we use it here to avoid clutter.

To finish the proof of this corollary, we need to show that any $Q$-function that satisfies the self-consistency equation in \cref{eq:Q_recursion} (which is what \cref{alg:FQE_earliest disagreement} seeks to produce) must also satisfy $Q(\gH_t) = \E_{P}[Y_{>t} \vert \gH_t]$ for every measurable $\gH_t$. We first derive the self-consistency condition from \cref{eq:expectation_identity}. Then we show that the only function that satisfies this condition is the conditional expectation we wish to estimate, $\E_{P}[Y_{>t} \vert \gH_t]$. 
\Cref{eq:expectation_identity} is rewritten below,
\begin{align*}
\E_{P}\left[ Y \vert \gH_{t} \right] &= 
\E_{\tgH \sim \tPr(\cdot \vert \gH_{t})}\left[\E_{P}\left[ Y ~\Big \vert ~ \gH_{t+\delta_{\tgH}(t)}= \gH_{t} \cup \tgH^{\setminus a_{\obs}}_{\left(t,t+\delta_{\tgH}(t)\right]} \right]\right].
\end{align*}
We write $Y = \sum_k{Y_k}$ as the sum of rewards that have been observed in the trajectory and a random variable that is the sum of future rewards.
\begin{align*}
\E_{P}\left[ Y \vert \gH_{t} \right] =
\E_{\tgH \sim \tPr(\cdot \vert \gH_{t})}\Bigg[\E_{P}\Bigg[ Y_{>t+\delta_{\tgH}(t)} + \!\! \sum_{(t_k,y_k)\in{\gH^y_{t+\delta_{\tgH}(t)}}}{y_k} ~\Big \vert ~ \gH_{t+\delta_{\tgH}(t)}= \gH_{t} \cup \tgH^{\setminus a_{\obs}}_{\left(t,t+\delta_{\tgH}(t)\right]} \Bigg]\Bigg]
\end{align*}
Then we subtract the rewards until time $t$, $\sum_{(t_k, y_k)\in{\gH^y_t}}{y_k}$ from both sides of the equation.
\begin{align*}
\!\!\E_{P}\left[ Y \vert \gH_{t} \right] - \sum_{(t_k, y_k)\in{\gH_t^y}}{\!\!\!\!\! y_k} = \E_{\tgH \vert \gH_{t} }\Big[ \!\!\!\!\!\!\!\!\!\! \sum_{\substack{(t_k, y_k)\in{\tgH^y}: \\ t_k\in{( t, t+\delta_{\tgH}(t)} ]}}{\!\!\!\! \! \!\! y_k} + \E_{P}\Big[ Y_{>t+\delta_{\tgH}(t)} \! ~\Big \vert \! ~ \gH_{t+\delta_{\tgH}(t)}= \gH_{t} \cup \tgH^{\setminus a_{\obs}}_{\left(t,t+\delta_{\tgH}(t)\right]} \Big]\Big]
\end{align*}
This is exactly \cref{eq:Q_recursion}, which we can simplify slightly to
\begin{align*}
\!\!\E_{P}\left[ Y_{>t} \vert \gH_{t} \right] = \E_{\tgH \vert \gH_{t} }\Big[ \sum_{\substack{(t_k, y_k)\in{\tgH^y}: \\ t_k\in{( t, t+\delta_{\tgH}(t)} ]}}{\!\!\!\! \! \!\! y_k} + \E_{P}\Big[ Y_{>t+\delta_{\tgH}(t)} \! ~\Big \vert \! ~ \gH_{t+\delta_{\tgH}(t)}= \gH_{t} \cup \tgH^{\setminus a_{\obs}}_{\left(t,t+\delta_{\tgH}(t)\right]} \Big]\Big]
\end{align*}
Next we show that if there is a unique function $Q(\cdot)$ that satisfies both: (1) $Q(\gH_T) = \E_{P}{[Y \vert \gH_T]}$, i.e. the estimator returns the correct outcome when it is given a full trajectory (notice the outcome is deterministic in this case), and (2) $Q(\cdot)$ satisfies the recursion in the above equation, namely
\begin{align} \label{eq:Q_recursion2}
    \!\!Q(\gH_t) = \E_{\tgH \vert \gH_{t} }\Big[ \sum_{\substack{(t_k, y_k)\in{\tgH^y}: \\ t_k\in{( t, t+\delta_{\tgH}(t)} ]}}{\!\!\!\! \! \!\! y_k} + Q(\gH_{t} \cup \tgH^{\setminus a_{\obs}}_{\left(t,t+\delta_{\tgH}(t)\right]}) \Big].
\end{align}
Due to this uniqueness, we will gather that $Q(\gH_t)$ must coincide with $\E_P{[Y_{>t} \vert \gH_t]}$ for all measurable $\gH_t$ and conclude the proof. 
\begin{lemma}
Assume $Q_1, Q_2$ are functions that satisfy both \cref{eq:Q_recursion2} and $Q_1(\tgH_T) = Q_2(\tgH_T) = \E[Y \vert \tgH_T]$ for all $\gH_T$. Then $Q_1(\gH_t) = Q_2(\gH_t)$ for all measurable $\gH_t$.
\end{lemma}
\begin{proof}
Since both $Q_1, Q_2$ satisfy \cref{eq:Q_recursion2}, it holds that
\begin{align*}
Q_1(\gH_t) - Q_2(\gH_t) = \E_{\tgH \vert \gH_t}[Q_1(\gH_{t} \cup \tgH^{\setminus a_{\obs}}_{\left(t,t+\delta_{\tgH}(t)\right]}) - Q_2(\gH_{t} \cup \tgH^{\setminus a_{\obs}}_{\left(t,t+\delta_{\tgH}(t)\right]})]
\end{align*}
Applying \cref{eq:Q_recursion2} repeatedly to $Q_1(\gH_{t} \cup \tgH^{\setminus a_{\obs}}_{\left(t,t+\delta_{\tgH}(t)\right]}) - Q_2(\gH_{t} \cup \tgH^{\setminus a_{\obs}}_{\left(t,t+\delta_{\tgH}(t)\right]})$ and so on, for say $d$ times, we get that
\begin{align*}
Q_1(\gH_t) - Q_2(\gH_t) = &\E_{\tgH \vert \gH_t}[ \E_{\tgH \vert \gH_{t+\delta_{\tgH}(t)}=\gH_t \cup \tgH^{\setminus a_{\obs}}_{\left(t,t+\delta_{\tgH}(t)\right]}}[ \\
&\E_{\tgH \vert \gH_{t+\delta_{\tgH}(t)+\delta_{\tgH}(t+\delta_{\tgH}(t))}=\gH_t \cup \tgH^{\setminus a_{\obs}}_{\left(t,t+\delta_{\tgH}(t)\right]} \cup \tgH^{\setminus a_{\obs}}_{\left(t+\delta_{\tgH}(t), t+\delta_{\tgH}(t)+\delta_{\tgH}(t+\delta_{\tgH}(t))\right]}}[ \ldots \\
&\qquad Q_1(\gH_{t} \cup \tgH^{\setminus a_{\obs}}_{\left(t,t+\delta^d_{\tgH}(t)\right]}) - Q_2(\gH_{t} \cup \tgH^{\setminus a_{\obs}}_{\left(t,t+\delta^d_{\tgH}(t)\right]})]]],
\end{align*}
where we denote the $d$-th disagreement under the nested sampling above by $\delta^d_{\tgH}(t)$. Notice that again there is slight abuse of notation here, since we reused the notation $\tgH$ in the expectations above, where the different $\tgH$ appearing above may not be identical trajectories. It holds that $\lim_{d\rightarrow \infty}{\delta^d_{\tgH}(t)} = T$ with probability $1$, since the number of events in a trajectory is finite with probability $1$. Hence we have $\lim_{d\rightarrow \infty}{Q_1(\gH_t\cup \tgH^{\setminus a_{\obs}}_{\left( t, t+\delta^d_{\tgH}(t) \right]}) - Q_1(\gH_t\cup \tgH^{\setminus a_{\obs}}_{\left( t, t+\delta^d_{\tgH}(t) \right]})} = Q_1(\gH_t\cup \tgH^{\setminus a_{\obs}}_{\left( t, T \right]}) - Q_1(\gH_t\cup \tgH^{\setminus a_{\obs}}_{\left( t, T \right]})$, and may conclude that because $Q_1(\tgH_T) = Q_2(\tgH_T) = \E[Y \vert \tgH_T]$ for full trajectories $\tgH_T$, the functions $Q_1(\cdot)$ and $Q_2(\cdot)$ must coincide, 
\begin{align*}
Q_1(\gH_t) - Q_2(\gH_t) = &\E_{\tgH \vert \gH_t}[ Q_1(\gH_{t} \cup \tgH^{\setminus a_{\obs}}_{\left(t,T\right]}) - Q_2(\gH_{t} \cup \tgH^{\setminus a_{\obs}}_{\left(t,T\right]})] = 0.
\end{align*}
\end{proof}
\end{document}

%% file: math_commands.tex
\usepackage{amsmath,amsfonts,bm}

\def\eqref#1{equation~\ref{#1}}

\def\1{\bm{1}}

\def\rva{{\mathbf{a}}}

\def\rve{{\mathbf{e}}}

\def\rvx{{\mathbf{x}}}
\def\rvy{{\mathbf{y}}}
\def\rvz{{\mathbf{z}}}

\def\rmA{{\mathbf{A}}}

\def\rmV{{\mathbf{V}}}

\def\rmX{{\mathbf{X}}}

\def\rmZ{{\mathbf{Z}}}

\def\vtheta{{\bm{\theta}}}

\DeclareMathAlphabet{\mathsfit}{\encodingdefault}{\sfdefault}{m}{sl}
\SetMathAlphabet{\mathsfit}{bold}{\encodingdefault}{\sfdefault}{bx}{n}

\def\gA{{\mathcal{A}}}

\def\gF{{\mathcal{F}}}

\def\gH{{\mathcal{H}}}

\def\gN{{\mathcal{N}}}

\def\gP{{\mathcal{P}}}

\def\gX{{\mathcal{X}}}

\def\sN{{\mathbb{N}}}

\def\sR{{\mathbb{R}}}

\newcommand{\E}{\mathbb{E}}



%% file: macros.tex
\usepackage{comment}
\usepackage{amsmath}
\usepackage{amssymb}
\usepackage{amsthm}
\usepackage{mathtools}
\usepackage{enumitem}
\usepackage{xspace}

\newtheorem{definition}{Definition}
\newtheorem*{definition*}{Definition}
\newtheorem*{lemma*}{Lemma}
\newtheorem*{theorem*}{Theorem}
\newtheorem{lemma}{Lemma}

\newtheorem{theorem}{Theorem}

\newtheorem{assumption}{Assumption}

\newtheorem{claim}{Claim}

\newcommand{\reals}{\mathbb{R}}

\DeclarePairedDelimiterX{\inp}[2]{\langle}{\rangle}{#1, #2}
\DeclarePairedDelimiterX{\abs}[1]{|}{|}{#1}
\DeclarePairedDelimiterX{\norm}[1]{\|}{\|}{#1}

\renewcommand{\eqref}[1]{Equation~(\ref{#1})}

\newlength\myindent 
\makeatletter
\newcommand*{\indep}{%
  \mathbin{%
    \mathpalette{\@indep}{}%
  }%
}
\newcommand*{\nindep}{%
  \mathbin{
    \mathpalette{\@indep}{\not}
  }%
}
\newcommand*{\@indep}[2]{%
  \sbox0{$#1\perp\m@th$}
  \sbox2{$#1=$}
  \sbox4{$#1\vcenter{}$}
  \rlap{\copy0}
  \dimen@=\dimexpr\ht2-\ht4-.2pt\relax
  \kern\dimen@
  {#2}%
  \kern\dimen@
  \copy0 
} 
\makeatother

%% file: paper_macros.tex
\newcommand{\obs}{\mathrm{obs}}
\newcommand{\rma}{\mathrm{a}}
\newcommand{\rmx}{\mathrm{x}}
\newcommand{\rmy}{\mathrm{y}}
\newcommand{\rmd}{\mathrm{d}}
\newcommand{\tgH}{\widetilde{\gH}}
\newcommand{\tPr}{\widetilde{P}}

\newcommand{\notcheckmark}{{\cmark}\textsuperscript{\textcolor{black}{\kern-0.75em{\bf---}}}}
\newcommand{\ours}{EDQ}

\newcommand{\dbtilde}[1]{\accentset{\approx}{#1}}

\usepackage{pifont}
\newcommand{\cmark}{\ding{51}}%
\newcommand{\xmark}{\ding{55}}%

\newcommand{\yw}[1]{{\color{blue}{YW: #1}}}

\renewcommand{\yw}[1]{\ignorespaces}

\newcommand{\attendto}[1]{{\color{cyan}{}}} 

%% file: sections/1_introduction.tex
\section{Introduction}
Sequential decision-making is common in 
healthcare, finance, and beyond \citep{chen2021probabilistic, upadhyay2018deep}. In hospitals, medical professionals administer treatments at different times based on the evolving observations of a patient's condition; in financial markets, traders execute orders based on sequential information flows. Algorithmic decision support systems can optimize these processes by evaluating different \textit{policies} with respect to their expected outcomes. 
%
%
Estimating the difference in expected outcomes between various policies is a causal effect estimation question~\citep{joshi2025ai, van2024algorithms,chen2021probabilistic}. This question involves several future treatment decisions taken at varying time points, hence it is a sequential decision-making problem. 


Formally,
this problem falls within the framework of off-policy evaluation \citep{uehara2022review, fu2021benchmarks}. A defining feature is that timings of observations and treatments are irregularly spaced, 
represented
by a stochastic point process with intensity $\lambda$, whereas the \emph{type} of the treatments at those times are specified as the marks of this process, governed by a distribution $\pi$. 
Unlike traditional formulations of off-policy evaluation that focus on action types, here the times must  
be accounted for, as they have a large effect on the outcome. This formulation is relevant to many decision-making scenarios, for example, in transplants, where it is often desirable to delay treatment as much as possible to lower the risk of complications. However, delaying too much can result in a deterioration in the patient's condition. Here, the type of treatment is fixed and what matters is the timing.


\textit{Estimating the effect of intervention on treatment timing} is a crucial part of evaluating sequential policies. With irregular times, frameworks for sequential decision-making that discretize time can be problematic, as 
discretization can be inaccurate or inefficient and requires choosing appropriate time scales. Further, length scales for decision-making within a single trajectory can vary dramatically. Consider for example a patient with heart failure; such a patient may be stable for months or years with occasional treatment adjustments made during 
visits to the cardiologist. However, at some point, they may experience acute decompensation~\citep{felker2011diuretic}, which requires rapid treatment and intensive monitoring during hospitalization.
Existing methods for continuous-time causal inference do not scale gracefully, since they solve complex estimation problems, such as integrating importance weights across time \citep{roysland2011martingale}. Those methods that scale to high-capacity models and large datasets do not handle dynamic policies (i.e., policies that take past states into account) and are implemented with differential equation solvers \citep{seedat2022continuous}, restricting architectural choices.

\begin{figure}
    
    \centering
    \includegraphics[width=0.85\linewidth]{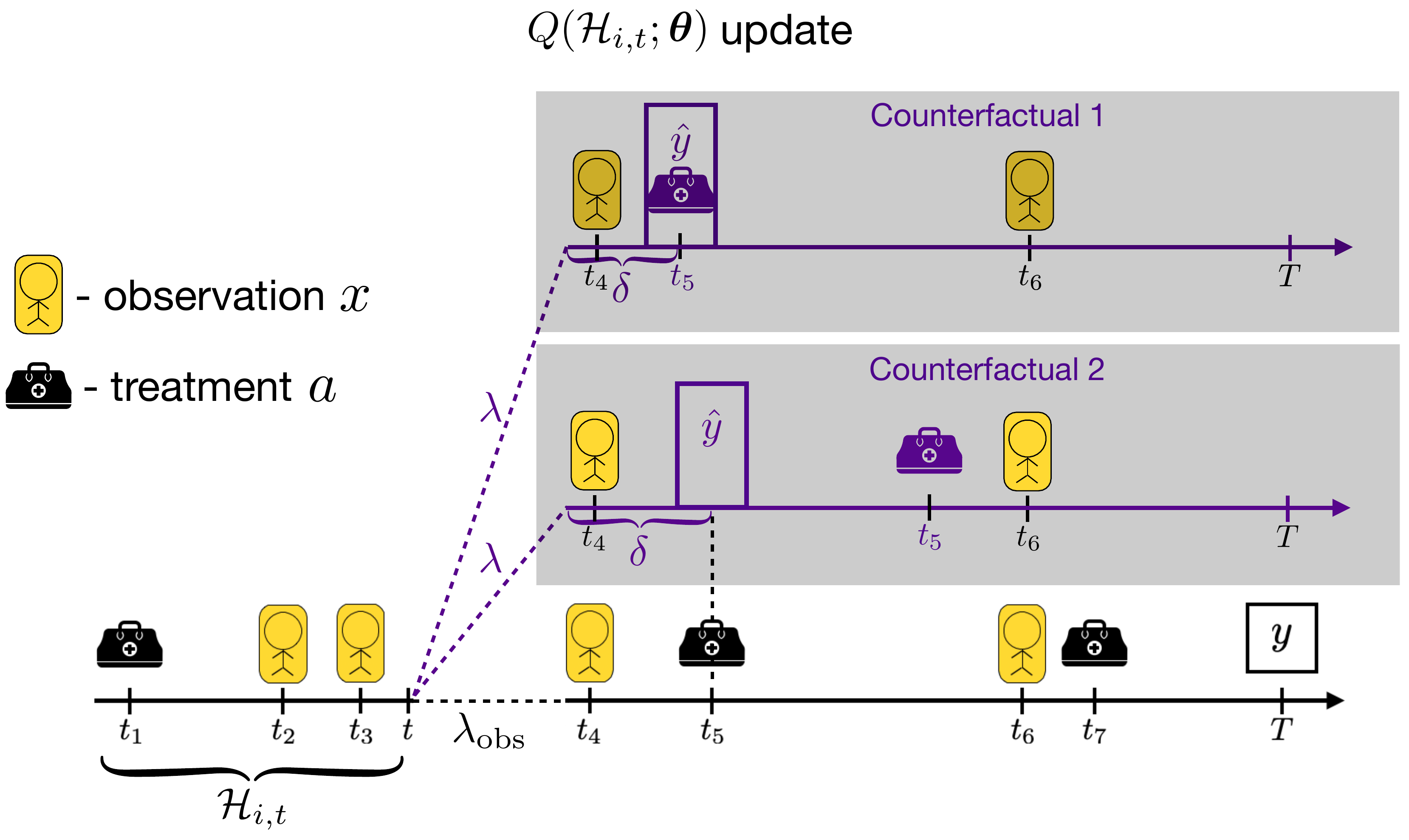}
    \caption{A summary of \ours{}. Conditional expectations of the outcome, $\E_{P}[Y \vert \gH_t]$, are estimated by
    $Q_t(\gH_t; \boldsymbol{\theta})$. Given an observed trajectory $\gH_{i}$ sampled under the training policy $\lambda_{\obs}$ (bottom trajectory), we fit values $Q_t(\gH_{i,t}; \boldsymbol{\theta})$ by regressing them on a label $\hat{y} := Q_{t+\delta}(\tgH_{i, t+{\delta}})$
    determined by a ``counterfactual" trajectory $\tgH$ sampled from the target policy $\lambda$. $\delta$ is the earliest disagreement time between the observed and counterfactual trajectories. It is either the time of the next observed treatment (counterfactual $1$) or that of the treatment under the target 
    policy (counterfactual $2$).
    }
    \label{fig:edq}
\end{figure}

In this work, we give two methods for off-policy evaluation with irregularly sampled data. Our contributions are as follows:
\begin{itemize}
    \item We define off-policy evaluation with \textit{decision point processes} and develop Earliest Disagreement Q-Evaluation (\ours{}), a model-free solution to the problem. While other methods are intractable in high dimensions or are limited to static treatments, \ours{} eliminates these restrictions. \ours{} is based on direct regressions and dynamic programming, which makes it easily applicable to flexible architectures including sequence models such as transformers. 
    
    \item In \Cref{thm:expectation_identity}, we prove that \ours{} is an empirical estimator of the correct policy value. The estimator produces accurate causal effects under assumptions on causal validity based on \citet{roysland2011martingale, roysland2022graphical}.
    \item We validate the efficacy of \ours{} through an experimental demonstration on time-to-failure prediction and tumor growth simulation tasks. For these tasks, we implement a transformer-based solution. The results show EDQ's advantage
    relative to baselines that rely on discretization.
\end{itemize}
We define the 
estimation problem, develop a solution, discuss related work and validate empirically.

%% file: sections/2_setting.tex
\section{Off-Policy Evaluation with Decision Point Processes} \label{sec:setting}
Consider a decision process defined by a marked point process $P$ \citep{andersen2012statistical, snyder2012random}
over observations (which take values in $\gX$), treatments (in $\gA$ respectively) and real outcomes. We are interested in estimating an overall quantity $Y \in \mathbb{R}$ that is a function of a sequence $Y_1, Y_2 \ldots$ of observed rewards. For convenience, we let $Y = \sum_{k=1}^{\infty}{Y_k}$, where $k$ is an index
for observed outcomes along the trajectory.\footnote{We define $Y_k(\omega) = 0$ for an event $\omega$ where $k$ is larger than the number of outcomes in the trajectory.} Though, the methods extend to other outcome functions like discounted future outcomes. We assume the number of rewards in the segment $[0,T]$ is finite.

\paragraph{Marked point processes.}
A marked point process is a
distribution over event times, along with distributions over \textit{marks}, or details of the events at each time (i.e. treatment times and which treatment was given). We consider multivariate counting processes $N(t) = (N^a(t), N^b(t) ,\ldots)$ on the time interval $[0,T]$. For a univariate process, e.g. $N^b$, $N^b(t)$ is the number of events of type $b$ until time $t$.
A trajectory of event times and their marks is a set $\gH = \{(t_0, \rve_0), (t_1, \rve_1), \ldots, (t_n, \rve_n)\}$. We denote events up to time $t$ by $\gH_t = \{(t_k, \rve_k)\in{\gH} : t_k \leq t \}$ and $\gH_{(t, t+\delta]}$ for events in the interval $(t, t+\delta]$. We use $\gH^{b}$ to refer to events of type $b$ on the trajectory and $\gH^{\setminus b}$ for events of all types other than $b$.

Finally, we assume intensity functions $\lambda(t \vert \gH_t) = \E[dN(t) \vert \gH_t]$ for processes exist, $N(t)$ is almost surely finite for any $t\in{[0,T]}$, and that the process can depend on its own history. That is, the filtration is the $\sigma$-algebra generated by random variables $N(t)$ and their marks \citep{aalen2008survival}.

\subsection{Problem Definition}
We follow notation from the RL literature \citep{upadhyay2018deep}. We begin with the data generating process and then summarize our goal of inferring causal effects.
This involves off-policy evaluation under a distribution $P$, while observing samples from $P_{\obs}$.

\begin{definition} \label{def:decpp}
A marked decision point process $P$ is a marked point process with observed components $N^e$ for $e\in{\{x, y, a\}}$ that have corresponding intensity functions $\lambda^e$, and mark spaces $\gX, \sR, \gA$, and a multivariate unobserved process with intensity $\lambda^u$. By default, we omit unobserved events from the trajectories $\gH$, hence $\gH = \{(t_0, \rve_0), (t_1, \rve_1), \ldots, (t_n, \rve_n)\}$ where $\rve_k\in\{ \gX \cup \gA \cup \sR \}$.
The intensity function and mark distribution $\lambda^a(t \vert \gH_t), \pi(A_t \vert \gH_t)$ are called the policy. The mark distributions for $X,Y$ are denoted by $P_X(\rvx_t \vert \gH_{t}), P_Y(y_t \vert \gH_{t})$.
\end{definition}

\paragraph{Off-policy evaluation.} We are given a dataset of $m$ trajectories, where trajectory $\gH_i$ has $n_i$ observations: $\{(t_{i,k}, \rve_{i,k})\}_{i\in{[m]}, k\in{[n_i]}}$.
These are sampled from an observed decision process $P_\obs$ with policy $(\lambda^a_{\obs}, \pi_{\obs})$.
Treatment times are samples from a counting process with intensity $\lambda^a_{\obs}$ and treatments at those times are sampled from $\pi_{\obs}$. We reason about outcomes when $(\lambda^a_{\obs}, \pi_\obs)$ is replaced with a target policy $(\lambda^a, \pi$), with other processes in $P_\obs$ fixed. The resulting decision process is denoted $P$, and our goal is to estimate the expected future outcome $\E_{P}\left[ Y \vert \gH_t \right]$ 
for all $t\in{[0,T)}$ and $\gH_t$ in the natural filtration associated with the point process.

\paragraph{When to treat.} To simplify notation, we omit the marks $\pi(A_t \vert \gH_t)$ and focus on intensities $\lambda^a(t \vert \gH_t)$. That is, we explore interventions on when to treat (\textit{medicine weekly or monthly?}) instead of how to treat (\textit{which medication?}). 
%
Technically, ``when" is the more challenging and underexplored part of the problem, and solutions can be easily extended to incorporate interventions on $\pi$ using existing methods \citep{chakraborty2014dynamic, pmlr-v158-li21a}. 
In the ASCVD example, this corresponds to reasoning about questions like: ``Consider prescribing statins to patients with characteristics $\gH_t$, whose LDL cholesterol is below 180 mg/dl up to time $t$. What would be the expected change in 10-year ASCVD risk if going forward, we prescribe a daily dose of statins for patients whose LDL cholesterol goes above 180 mg/dL, instead of the existing policy followed in the population?".


\subsection{Roadmap to identifiability via Local Independences}
\label{sec:identifiability}
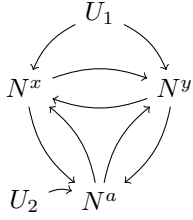
\begin{wrapfigure}[13]{l}{0.5\textwidth}
\centering
\begin{tikzpicture}
      \node (U) at (0, 1) {$U_1$};
      \node (Nx) at (-1, 0) {$N^x$};
      \node (Ny) at (1, 0) {$N^y$};
      \node (Na) at (0, -1.5) {$N^a$};
      \node (Ua) at (-1, -1.5) {$U_2$};
    
      \draw[->, bend right=20] (U) to (Nx);
      \draw[->, bend left=20] (U) to (Ny);
      \draw[->, bend left=20] (Nx) to (Ny);
      \draw[->, bend left=20] (Ny) to (Nx);
      \draw[->, bend right=20] (Nx) to (Na);
      \draw[->, bend right=20] (Na) to (Nx);
      \draw[->, bend left=20] (Na) to (Ny);
      \draw[->, bend left=20] (Ny) to (Na);
      \draw[->, bend left=20] (Ua) to (Na);
\end{tikzpicture}
\caption{The assumed local independence graph for a decision point processes, where our estimand is identifiable from observed data $(N^x, N^a, N^y)$.}
\label{fig:local_graph}
\end{wrapfigure}
The goal of this section is 
to elucidate the conditions under which the algorithm we present in \cref{sec:method} estimates valid causal effects. We briefly summarize the essential conditions and supplement this summary in \cref{app:identifiability}.
Our assumptions to ensure identifiability of causal estimands follow \citet{roysland2011martingale, roysland2022graphical, didelez2008graphical}, who study graphical models for point processes.
In this setting,
where the goal is to intervene on $N^a$ and estimate $E_P[Y \vert \gH_t]$ under $P$ rather than $P_{\obs}$, in the presence of unobserved processes $U$,
\cite{roysland2012counterfactual} define and analyze the following notions:\\
\begin{itemize}
    \item A graphical condition called \textit{causal validity} ensures that changing the treatment intensity 
    from $\lambda^a_{\obs}$ to the interventional $\lambda^a$, while changing no other intensities, changes the joint
    distribution from $P_\obs$ to $P$. A graph may not be causally valid 
    when it contains unobserved variables $U$. 
    \item \textit{Local independence }\citep{aalen1987dynamic, schweder1970composable} adapts sequential exchangeability \citep{robins1986new, hernan2023causal} to continuous time. It is an asymmetric form of independence where one process' intensity may depend on 
    the history of another, but not vice versa.

    \item A certain set of local independences 
    together are referred to as \textit{eliminability}  (a generalization of the backdoor criterion), which implies casual validity,
    even with unobserved variables $U$.
\end{itemize}
Consider the graph in \Cref{fig:local_graph}, where an edge means that the history of the source node affects the future of the target node \citep{didelez2008graphical}. It is possible to show
that it satisfies
eliminability
(see \cref{app:identifiability}).
To understand this condition, we start with the basic local independence requirements.
\begin{definition}
For a multivariate process $N(t)=(N^a(t),N^b(t),\ldots)$ on variables $V$ we say that $N^a$ is locally independent of $N^b$ given $N^{\setminus b}$, or $N^b \centernot\rightarrow N^a \vert N^{\setminus b}$, if
$\lambda^a(t \vert \gH_{t})=\lambda^a(t \vert \gH^{\setminus{b}}_{t})$.
A graphical local independence model $(\gP, G)$ is a class of processes $\gP$ on $V$ and directed graph $G=(V, E)$, such that $(b\rightarrow a) \notin {E}\Rightarrow N^b \centernot\rightarrow N^a \vert N^{\setminus b}$ holds for all $P\in{\gP}$.
\end{definition}
This condition means that the intensity of a process only makes use of certain information from other processes.
\cite{roysland2022graphical} 
package together the set of local independences
that imply causal validity under the name \textit{eliminability}, defined below. They use a graphical criterion that is akin to using d-separation in graphical models, while we state the conditions in terms of the implied functional independencies of intensities. We expand on this in \cref{app:identifiability}.
\begin{definition}
Let $U$ be the set of unobserved variables.
   Suppose $U$ can be written as a sequence $(U_1, \ldots, U_K)$ such that for each $k$, either
\begin{itemize}
    \item $(N^y, N^x, U_{>k})$ is locally independent of $U_k$ given $(N^x, N^y, N^a, U_{>k})$, or 
    \item $N^a$ is locally independent of $U_k$ given $(N^x, N^y, N^a, U_{>k})$.
\end{itemize}
Then the graph is said to satisfy Eliminability. Here, for sets $A, C$ and variable $b$ we say $A$ is locally independent of $b$ given $A\cup C$ if each $a\in{A}$ is locally independent of $b$ given $A\cup C \setminus \{a\}$.
\label{def:elim}
\end{definition}

We summarize that \textit{causal validity} 
holds under the \textit{local independences}
that satisfy \textit{eliminability}.
We assume the graph in \Cref{fig:local_graph}, which satisfies these assumptions. We
also assume mutual independence
of increments of all processes at a one time to rule out instantaneous effects. We refer to
validity and this independence
together as \textbf{ignorability}, in accordance with existing terminology on confounding.
\begin{assumption}
\label{assump:ignor}
\textit{Ignorability (in continuous time)} is satisfied when:
   \begin{enumerate}
       \item the graph satisfies causal validity,
       \item the increments
   of features, treatments, and outcome are mutually independent given
the history, i.e.,
$((dN^x(t),  X_t) \indep (dN^a(t), A_t) \indep (dN^y(t), Y_t))    | \mathcal{H}_t$.
   \end{enumerate}
\end{assumption}
In addition to ignorability, we require a second, standard assumption, \textbf{overlap}, for the conditional expectations we  estimate to be well-defined.
Recall that the interventional distribution $P$ 
is defined by replacing 
the treatment distribution in $P_{\obs}$, i.e., 
replacing
$\lambda_{\obs}^a$ with $\lambda^a$.
\begin{assumption}
    Overlap is said to hold between the observational and interventional distributions, $P_{\obs}$ and $P$,
   if $P$ is absolutely continuous with respect to $P_{\obs}$, denoted by $P \ll P_{\obs}$.
    \label{assump:overlap}
\end{assumption}

Ignorability and overlap are the core assumptions that allow identification in our setting. 
Under these assumptions, we can 
now present algorithms for estimating causal effects in continuous time.

%% file: sections/4_estimation.tex
\section{Model Free Off-Policy Evaluation for Decision Point Processes} \label{sec:method}
To estimate $\E_{P}[ Y \vert \gH_t ]$ for times $t$ and $\gH_t$ that overlap with $P_\obs$, we express the expectation recursively as a function of expectations $\E_{P}[Y \vert \tgH_{t+\delta} ]$ for some $\delta > 0$ and trajectory $\tgH$. Then, assuming that expectations at times larger than $t$ have been learned correctly, this recursive expression allows us to propagate information for conditioning on earlier histories. Let us describe this solution in more detail, as applied in the context of $Q$-evaluation. 
\begin{wrapfigure}{r}{0.5\textwidth}
    \input{Figures/FQE_vanilla}
\end{wrapfigure}

\textbf{Fitted Q evaluation (FQE) in discrete time.} Q-evaluation relies on the tower property of conditional expectations, given below in \cref{eq:tower}. In discrete time decision processes, where we consider $\delta=1$, the property suggests a dynamic programming solution that we lay out in \cref{alg:FQE} \citep{le2019batch, watkins1992q}. Here, $\gH_t$ includes all treatments and observations up to and including time $t$, and 
since they occur simultaneously, there are exactly $t$ of each. 
$\tgH_{t+1}$ is defined in the same manner, except that it includes $\tilde{a}_{t+1}$ sampled from the target policy $\pi$.
\begin{align} \label{eq:tower}
\E_{P}\left[Y \vert \gH_t \right] = \E_{\tgH_{t+\delta}\sim P(\cdot \vert \gH_t)}\left[\E_{P}\left[ Y \vert \tgH_{t+\delta} \right] \right],
\end{align}
An attractive property of this algorithm is that it is \emph{model-free}. That is, to form the label $\hat{y}_i$ we only need to sample $\tilde{a}_{t+1}$ from our target policy $\pi$, while a model of $P(\rvx_{t+1} \vert \gH_t)$ is not necessary.

With accurate optimization over a sufficiently expressive hypothesis class and arbitrarily large datasets, \cref{alg:FQE} returns correct estimates.
This is because if we fix $Q_{t+1}(\gH_{t+1})$ and assume that it accurately estimates $\E_{P}[ \sum_{s\geq t+1} Y_s \vert \gH_{t+1} ]$, then the minimizer of the regression is the conditional expectation, equal to 
$\E_{P}[ \sum_{s\geq t} Y_s \vert \gH_{t} ]$ according to \cref{eq:tower}.
The model-free solution is enabled by the equality $P(\rvx_{t+1} \vert \gH_t) = P_{\obs}(\rvx_{t+1} \vert \gH_t)$, which validates the use of $\rvx_{i, t+1}$ to form $\hat{y}_i$. 
In practice, we take gradient steps on randomly drawn times and training samples instead of walking backward from $T$ to $1$.
Crucially, for $\delta > 1$, e.g. $\delta=2$, we have $P(\rvx_{t+2} \vert \gH_t) \neq P_{\obs}(\rvx_{t+2}\vert \gH_t)$. Hence, an algorithm using \cref{eq:tower}
must
either be model-based, or resort to solutions such as 
importance weights that suffer high variance \citep{precup2000eligibility, hallak2016generalized}, or restrict the problem, e.g., by discounting rewards \citep{munos2016safe, harutyunyan2016q, precup2000eligibility}.

\textbf{Challenges in application to continuous time.} Moving to continuous time, the tower property turns into a differential equation, and solving it requires tools that go beyond common FQE (e.g., \citet{jia2023q}). Regressing to an outcome that is arbitrarily close to the observation at time $t$ is ill-defined. While we may work under a fine discretization of time, this approach is wasteful, as a single update in the minimization for estimating $Q_t$ takes into account the development of the process in the interval $[t, t+\delta]$, and for small values of $\delta$ this will usually yield a very small change to the estimate. Hence, intuitively, when updating $Q_t$, we would like to use estimates of $Q_{t+\delta}$ for a large $\delta$. As explained above, this is seemingly difficult to achieve in a model-free fashion. However, for point processes, since the number of decisions over $[0,T]$ is countable, it seems plausible that a simple and efficient dynamic programming solution can be devised. In what follows, this is what we present.

\subsection{\ours{}: Fitted Q-Evaluation for Decision Point Processes via Earliest Disagreement Times} \label{sec:edq}

We wish to reason about what the outcome would have been for an observed trajectory sampled under policy $\lambda^a_{\obs}$, had we treated it with policy $\lambda^a$ from time $t$ onward. Intuitively, it seems plausible that we can use a similar approach to FQE, but instead of going one time unit forward, we can move to the first time $t+\delta$ where the two policies differ in their treatments. That is, we will sample alternative treatment trajectories, asking what the target policy $\lambda^a$ would have done at each time, given the observed history up until that time and find the earliest disagreement. The resulting algorithm is rather simple, and we summarize it graphically in \cref{fig:edq}. The attractive property of this approach is that the ``lookahead" time $\delta$ is adaptive. It will likely be short when applied in parts of trajectories where many treatments are applied, and longer when treatments are scarce.

To formalize the method, we present some additional notation, prove an appropriate variation of the tower property, and explain how it is operationalized by the implementation in \cref{alg:FQE_earliest disagreement}. 

\begin{definition} \label{def:future_dist}

For process $P_{\obs}$ and policy $(\lambda^a, \pi)$, define an augmented process $\tilde{N} = (N^{a_{\obs}}, N^y, N^x, N^{a})$ where intensities are independent of the history of $N^{a_{\obs}}$: (1) the intensity of $N^{a}$ is $\lambda^{a}(t \vert \gH^{\setminus a}_t) := \lambda^a(t \vert \gH_t^{x,y}, \gH_t^a=\gH_t^{a_{\obs}})$, i.e. $\lambda^a$ where history of treatments is given by $\gH_t^{a_{\obs}}$,
\footnote{Note that this is a slight abuse of notation, since $\gH_t^a$ is not a random variable.}
and (2) $\lambda^{e}(t \vert \gH_t) = \lambda^{e}_{\obs}(t \vert \gH_t^{\setminus a})$ for $e\in{\{a_{\obs}, x, y\}}$. 
For a trajectory $\tgH\sim\tilde{P}$ and time $t\in{[0,T)}$, define $\delta_{\tgH}(t) = \min\{u-t : u > t, (u,\cdot)\in{\tgH^{a, a_{\obs}}}\}$,
where $\delta_{\tgH}(t)=T-t$ when the set
is empty.




\end{definition}
The augmented process maintains an additional treatment trajectory, $\tgH^a$, as an alternative to $\tgH^{a_{\obs}}$, the one observed under $\lambda_{\obs}$. From \cref{def:future_dist}, 
we observe that the intensities of the augmented process $\tilde{N}$
do not depend on $N^{a}$'s history. It follows that the marginal over $\{x,a,y\}$ is $P_{\obs}$, while $\tgH^{a}$ has a similar role to the alternative treatment $\tilde{a}_{t+1}$ in \cref{alg:FQE}. This notation is helpful for denoting sampled alternative trajectories over time intervals. Finally, $\delta_{\tgH}(t)$ denotes the earliest disagreement between observed and target treatments after time $t$.\footnote{The minimum treatment time is also the earliest disagreement, since under some regularity conditions, the processes $N^a$ and $N^{a_{\obs}}$ have probability $0$ of jumping simultaneously.}
Our model-free evaluation method is based on the following result, which expresses our estimand as an expectation over trajectories $\tgH$.

\begin{restatable}{thm}{expectationid}
\label{thm:expectation_identity}
Let $P_\obs$ be a marked decision point processes, $P$ the process obtained by replacing the policy with $(\lambda^a, \pi)$, and $\tilde{P}$ the augmented process obtained from $P,P_{\obs}$ in \cref{def:future_dist}. Further, let $t\in{[0, T)}$, and $\gH_t$ measurable w.r.t $P$.
Under Assumption~\ref{assump:overlap}, 
we have that
\begin{align} \label{eq:expectation_identity}
\E_{P}\left[ Y \vert \gH_{t} \right] = 
\E_{\tgH \sim \tPr(\cdot \vert \gH_{t})}\left[\E_{P}\left[ Y ~\Big \vert ~ \gH_{t+\delta_{\tgH}(t)}= \gH_{t} \cup \tgH^{\setminus a_{\obs}}_{\left(t,t+\delta_{\tgH}(t)\right]} \right]\right].
\end{align}
\end{restatable}
\textbf{Takeaways from \Cref{thm:expectation_identity} and derivation of \ours{}.}  
\Cref{eq:expectation_identity} suggests a method to calculate expectations $\E_{P}[Y \vert \gH_t]$, similar to how FQE follows from \cref{eq:tower}.
The practical version of the resulting method, \ours{}, is given in \cref{alg:FQE_earliest disagreement}. 
To arrive at the method from \cref{eq:expectation_identity}, let us examine the regressions solved in \cref{alg:FQE_earliest disagreement}. Taking a random variable $Y_{>s}$ which is the sum of all outcomes after some time $s$, \cref{eq:expectation_identity} can be rewritten as follows (see \Cref{app:experiments_comments} for a detailed derivation).
\begin{align} \label{eq:Q_recursion}
\!\!\underbrace{\E_{P}\left[ Y \vert \gH_{t} \right] - \!\!\!\!\!\!\!\! \sum_{(t_k, y_k)\in{\gH_t^y}}{\!\!\!\!\! y_k}}_{Q(\gH_{t})} = \E_{\tgH \vert \gH_{t} }\Big[ \!\!\!\!\!\!\!\!\!\! \sum_{\substack{(t_k, y_k)\in{\tgH^y}: \\ t_k\in{( t, t+\delta_{\tgH}(t)} ]}}{\!\!\!\! \! \!\! y_k} + \underbrace{\E_{P}\Big[ Y_{>t+\delta_{\tgH}(t)} \! ~\Big \vert \! ~ \gH_{t+\delta_{\tgH}(t)}= \gH_{t} \cup \tgH^{\setminus a_{\obs}}_{\left(t,t+\delta_{\tgH}(t)\right]} \Big]}_{Q(\gH_{ t+\delta_{\tgH}(t)})}\Big]
\end{align}
\Cref{alg:FQE_earliest disagreement} uses gradient descent to fit $Q$ functions that, at optimality, satisfy a self-consistency condition that appears in the equation above, $Q(\gH_t) = \E_{\tgH}[\sum{y_k} + Q(\gH_{t+\delta_{\tgH}(t)})]$. To this end, it solves regression problems where $Q(\gH_{i,t})$ is fitted to a label $\hat{y}_i$. This label is defined in line $7$ and coincides with the term in the expectation $\E_{\tgH \vert \gH_t}[\cdot]$ above. To conclude that the algorithm estimates the effect of interest, two simple arguments suffice: (1) a uniqueness argument showing that the self-consistency equation holds only when $Q(\gH_t)$ matches $\E_{P}[Y_{>t} \vert \gH_t]$, and (2) that despite ignoring the unobserved process $\lambda^u$, the estimation yields the desired causal effect due to the ignorability assumption. We summarize the conclusion below and provide detailed steps in \cref{app:identifiability}. 
\begin{restatable}{coro}{edqcorro}\label{corro:EDQ_estimate}
Under assumption~\ref{assump:ignor}, a $Q$-function satisfying \cref{eq:Q_recursion} yields the causal effect of the intervention that replaces $(\lambda^{a}_{\obs}, \pi_{\obs})$ with $(\lambda^a, \pi)$.
\end{restatable}

An analogue of \cref{eq:expectation_identity} holds for discrete-time decision processes, where 
\ours{} bears some resemblance to the eligibility traces approach of \citet{precup2000eligibility}. We provide this result in \cref{sec:tower_expectation_discrete} for completeness but focus here on the point process case, as this is our main motivation and where the earliest disagreement approach is most fruitful.
\begin{figure}
\input{Figures/ED_FQE}  
\end{figure}

%% file: Figures/FQE_vanilla.tex
\vspace{-10pt}
\begin{minipage}[t]{\linewidth}
\begin{algorithm}[H]
\caption{Fitted Q-Evaluation (discrete time)}
\label{alg:FQE}
\vspace{15pt}
\begin{algorithmic}[1]
    \State \textbf{Input}: Trajectories $\{\gH_i\}_{i=1}^{m}$, \\
    \qquad Policy $\pi:\cup_{t=1}^{T}\gX^{t}\times\gA^{t-1} \rightarrow \Delta^{|\gA|}$ \\
    \qquad Model class $\gF$ where $f:\cup_{t=1}^{T}\gX^{t}\times\gA^{t} \rightarrow \sR $

    \State Initialize $Q$ randomly
    \State Set $T_i \gets \max \{ t : (t,\rvz)\in{\gH_i}\} \quad \forall i\in{[m]}$
    \For{$N$ rounds}
        \State Draw $\tilde{a}_{t+1}\sim \pi(\cdot \vert \gH_{i,t}, \rvx_{i, t+1}) \quad \forall i, t\in{[T_i]}$
        \State $\tgH_{i,t+1} \gets \gH_{i,t} \cup \{(t+1, \rvx_{t+1}),(t+1, \tilde{a}_{t+1})\}$
        \State Set $\hat{y}_{i,t} = y_{i,t} +  Q(\tgH_{i,t+1}) \quad \forall i,t\in{[T_i]}$
        \State $\displaystyle Q \gets
\mathrm{arg}\min_{f\in{\gF}}\sum_{i\in{[m]}, t\in{[T_i]}}\left( f(\gH_{i,t}) - \hat{y}_{i,t} \right)^2$
    \EndFor
    
    \State \textbf{Return} $Q$
    
\end{algorithmic}
\end{algorithm}
\end{minipage}

%% file: Figures/ED_FQE.tex
\vspace{-15pt}
\begin{minipage}[t]{\linewidth}
\begin{algorithm}[H]
\caption{Earliest Disagreement Fitted Q-Evaluation}
\label{alg:FQE_earliest disagreement}

\begin{algorithmic}[1]
    \State \textbf{Input}: Trajectories $\{\gH_i\}_{i=1}^{m}$, \\
    \qquad\quad Policy $\lambda_a(\cdot \vert \gH_t), \pi(\cdot \vert \gH_t)$

    \State Initialize $\vtheta$ randomly
    \For{ $N$ rounds}
        \State Draw $t\sim \mathrm{Unif}([0,T])$ and $i \sim \mathrm{Unif}([m])$
        \State Draw $\tgH\sim \tPr(\cdot \vert \gH_{i, t})$ and set $\gH'_{i, t+\delta_{\tgH}(t)} = \gH_t \cup \tgH^{\setminus a_{\obs}}_{\left(t, t+\delta_{\tgH}(t)\right]}$ 
        \State $\hat{y}_{i} = \sum_{\substack{(t_k, y_k)\in{\tgH^y}: \\ t_k\in{( t, t+\delta_{\tgH}(t)} ]}} y_{k} + Q(\gH'_{i, t+\delta_{\tgH}(t)}; \vtheta)$ 
        \State $\vtheta \leftarrow \vtheta - \eta \nabla_{\vtheta} \left(Q(\gH_{i,t}; \vtheta) - \hat{y}_i\right)^2$
    \EndFor
    
    \State \textbf{Return} $Q(\cdot; \vtheta)$
    
\end{algorithmic}
\end{algorithm}
\end{minipage}
\vspace{-15pt}

%% file: sections/3_related_work.tex
\section{Related Work} \label{sec:related_work}

\vspace{-1mm}

Our coverage of related work is divided into an overview of works that solve adjacent tasks to ours, before transitioning into a detailed discussion in \cref{sec:related_work_b} about techniques more closely aligned with our goal of large scale causal inference in sequential decision making.

\textbf{Causal inference with sequential decisions.} Causal effect estimation for sequential treatments is usually studied in discrete time under the sequential exchangeability assumption \citep{robins1986new, hernan2023causal}. Addressing unobserved confounders is of interest \citep{tennenholtz2020off, namkoong2020off}, but this is beyond our scope here. As described in \cref{sec:identifiability}, the framework of \citet{roysland2022graphical} draws a parallel to sequential exchangeability for continuous time, which we adopt here. For estimation in these continuous-time problems, several methods have been explored \citep{lok2008statistical, lin2004analysis, rytgaard2022continuous, roysland2011martingale, zhang2011causal}. Most do not scale to large and high-dimensional datasets. For instance, \citet{roysland2011martingale} requires estimating an integral of propensity weights over time; \citet{rytgaard2022continuous} propose a targeted estimator that fits each process in $P_{\obs}$, but it 
is limited to interventions at fixed times (i.e. no interventions on treatment schedules) and it is unclear how to implement it with expressive models. \citet{nie2021learning} study the effect of timing $t$ when applying a fixed policy from time $t$ onward. However, they do not discuss interventions on all treatment timings, as we do here.

\textbf{Reinforcement learning (RL) techniques.} As discussed in \cref{sec:method}, \ours{} is related to FQE and Q-learning \citep{le2019batch, watkins1992q, murphy2005generalization}, and to $n$-step methods from RL \citep{munos2016safe, de2018multi, precup2000eligibility}. In the causality literature, Q-learning often appears in the context of dynamic treatment regimes \citep{chakraborty2014dynamic, chakraborty2013statistical}, where discrete-time policies are learned off-policy. All these methods are not applicable to irregularly sampled times.
Works in RL that consider irregularly sampled times, either do not intervene on treatment times, or operate in the on-policy setting \citep{qu2023bellman, upadhyay2018deep}, or incorporate continuous-time positional embeddings into decision transformers \citep{chen2021decision}. The latter facilitates the recommendation of sets of actions to arrive at a desired outcome \citep{zhang2023continuous} rather than evaluating a policy of interest. 
Recent work suggests that goal-conditioned imitation learning methods such as decision transformers may fail to estimate the causal effect of actions \citep{malenica2023causality} in some scenarios where there are no unobserved confounders, whereas $Q$-learning methods produce correct estimates.

We next discuss \emph{scalable} causal estimation methods for sequential treatments. We delineate 
assumptions, implementation choices, and subsequent properties of solutions from recent work.

%% file: sections/3B_related_work_technical.tex
\subsection{Large Scale Estimation Approaches for Sequential Treatments} \label{sec:related_work_b}
\input{Figures/comparison_table}
Notable early work on machine learning for estimating counterfactual quantities related to treatment timelines,
\citet{schulam2017reliable},  used Gaussian Processes for estimation. Limitations such as scalability 
and incorporating various features prompted the development of deep learning approaches.

\textbf{Large scale models.} One family of solutions \citep{lim2018forecasting, Bica2020Estimating, melnychuk2022causal} took an important step forward by using RNNs and transformers for estimation. All of these methods build on the idea of learning balancing representations \citep{johansson2016learning}. Roughly, these are representations under which the treatment is randomly assigned. This facilitates the training of high-capacity effect estimators on large datasets, but these works are restricted to discrete times.

\textbf{Dynamic policies.} The above methods can only estimate effects of \emph{static} treatments, meaning the treatment plan cannot dynamically depend on future observations. For instance, consider a policy that prescribes a daily dose of statins, and if the patient develops side effects in the future, switches to another medication. This policy depends on possible future states, which the above methods do not accommodate. G-Net \citep{pmlr-v158-li21a, xiong2024g} takes a model-based approach, where models are fit for both $\pi_{\obs}(A(t) \vert \gH_{t-1}, X(t))$, and $P_{\obs}(X(t), Y(t) \vert \gH_{t-1})$. Then at inference time, a dynamic policy is estimated when $\pi_{\obs}$ is replaced with the desired policy $\pi$ and conditional expectations of $Y$ are estimated with Monte-Carlo simulations.

\textbf{Irregular times.} None of the above solutions address irregular observation and action times, a key focus of this paper. Some steps in this direction have been taken by \citet{seedat2022continuous, vanderschueren2023accounting}. 
Past work on irregular times includes \citet{seedat2022continuous}, who combine balanced representations with a neural controlled differential equation architecture suited for irregular sampling, while \citet{vanderschueren2023accounting} use reweighting to adjust for sampling times that are informative of the outcome. However, these methods mitigate sampling-induced bias rather than estimate outcomes under interventions on treatment times. 
We further discuss them in \cref{app:related_work}.

\Cref{tbl:related_methods} summarizes the 
above techniques, along with 
FQE
and \ours{}.
Notably, \ours{} possesses the desirable qualities mentioned here and handles interventions on $\lambda^a$, which other solutions do not.

%% file: Figures/comparison_table.tex
\begin{table}[ht!]
\centering
\caption{Qualitative comparison of  estimation methods under sequential treatments, compatible with large scale ML. Our method EDQ is marked in bold, while others can be found in \citep{schulam2017reliable,Bica2020Estimating, melnychuk2022causal, lim2018forecasting, seedat2022continuous, pmlr-v158-li21a, chakraborty2013statistical, le2019batch}. 
The acronym DP stands for Dynamic Programming.}
\label{tbl:related_methods}
\begin{tabular}{lccc cccc}
\toprule
& \multicolumn{3}{c}{\textbf{Properties}} & \multicolumn{4}{c}{\textbf{Estimation Method}} \\
\cmidrule(lr){2-4} \cmidrule(lr){5-8}
& {\makecell{Irregular\\ Times}} & {\makecell{Large \\ Scale}} & {\makecell{Dynamic \\ Policy}} & {\makecell{Prop. \\ Weights}} & {\makecell{Model \\ Based} } & {\makecell{Balancing \\ Rep.}} &
{\makecell{DP}}
\\ 
CGP & \xmark & \xmark & \cmark & & \Large$\bullet$ & &
\\ 
CRN , CT & \xmark & \cmark & \xmark &  & & \Large$\bullet$ & 
\\ 
R-MSN & \xmark & \cmark & \xmark & \Large$\bullet$ & & & 
\\ 
TE-CDE & \cmark & \xmark{}\tablefootnote{We mark TE-CDE with as non-scalable, since the algorithm relies on differential equation solvers which limits the scalability of methods one can use.} & \xmark & & \Large$\bullet$ & \Large$\bullet$ & \\ 
 G-Net & \xmark & \cmark & \cmark & & \Large$\bullet$ & &\\  
FQE & \xmark & \cmark & \cmark & & & & \Large$\bullet$ \\ 
\textbf{EDQ} & \cmark & \cmark & \cmark & & & & \Large$\bullet$ \\ 
\bottomrule
\end{tabular}
\end{table}

%% file: sections/5_experiments.tex
\section{Implementation and Experiments} \label{sec:experiments}
To implement \ours{} for experimentation 
we use the GPT-2 architecture. 
Each token is a concatenation of embeddings of time $t_i$, value $\rvz_i$ and event type $e_i\in{\{A, X, Y, \Delta T\}}$. The event types $A,X,Y$ correspond to actions, features, and outcomes, while $\Delta T$ is introduced for convenience to represent
cases where time passes but no events occur. 
Both absolute times $t_i$ and time gaps $\Delta T$
use continuous time positional embeddings: the $k$-th dimension is $
\sin \left( t C^{-k /  d_{\mathrm{time}}} \right)$ for even $k$, and $\cos \left( tC^{-{(k-1)} / d_{\mathrm{time}}} \right)$ for odd $k$. Here $C=10^5$ 
and $d_{\mathrm{time}}$ is the embedding dimension. We also keep a target network and update it with soft-Q updates, as is common in Deep Q-Networks, e.g., \citet{van2016deep}.
\subsection{Baselines}
We are unaware of 
baselines for effect estimation on treatment timing with high-dimensional or long-sequence data. Thus, we implement two baselines that let us glean important aspects of \ours{}.

\textbf{ERM / MC} is an Empirical Risk Minimizer (ERM) trained to predict observed outcomes, which, also known as Monte-Carlo (MC) prediction in RL and policy evaluation.
We use the same GPT-2 architecture and data representation as \ours{}, but instead of running \cref{alg:FQE_earliest disagreement}, we train $f_{\vtheta}(\gH_t)$ by minimizing prediction loss on observed data. 
Given each training trajectory $\gH_i$ with an outcome label $y_i$, we solve $\min_{\theta} \sum_{i,(t_i, \rvz_i)\in{\gH_{i}}} (\ell(f_{\vtheta}(\gH_{i,t}), y_i)$
, where $\ell(\cdot, \cdot)$ is the squared loss.
This method estimates outcomes under $\lambda_{\obs}$, therefore we expect it to perform as well as or better than off-policy evaluation methods, including \ours{}, when $\lambda=\lambda_{\obs}$, and to suffer a drop otherwise.

\textbf{FQE} is implemented following \cref{sec:method}, but with discretized time and Q-updates using one timestep forward.  At each iteration, we draw training example $i\in{[m]}$ and time $t\in{[0,T]}$, define $\hat{y}_i = y_{t+1} + Q_{t+1}(\tgH_{t+1}; \vtheta)$), and take a gradient step on loss $\ell(Q_t(\gH_{i,t}), \hat{y}_i)$. Similarly, we define discrete-time approximations of our policies of interest, described later. The positional embeddings correspond to discrete times, and the representations of actions, features and outcomes at each time are concatenated. Other than this, we use the same architecture and hyperparameters of \ours{}. This baseline examines the effects of time discretization on estimation and optimization.

\emph{Computational complexity:} The per-iteration runtime of \ours{} is similar to FQE, which is a common tool for large-scale offline RL problems, e.g. \citet{paine2020hyperparameter, voloshin2021empirical}. 
\ours{} and FQE differ in computation times due to sampling methods from the target policy used to draw the treatments used in the $Q$-update. We discuss this 
in \cref{app:experiments_comments}.

\subsection{Simulations on Time to Failure and Cancer Tumor Growth Prediction}
To validate 
our method, we construct two settings. The first is to predict the effect of treatment timing policies on patients' time-to-event. The second uses a cancer tumor growth simulator from \citet{geng2017prediction} to form a policy evaluation problem on applications of chemotherapy and radiotherapy.

%
\textbf{Simulators.} We use two simulators. \textbf{(i) Time-to-failure:} In this setting, each data point simulates the vital of a patient $x_t\in{\reals_+}$ measured regularly at a frequency of one time unit, and treatments $a_t\in{\reals_+}$ are assigned irregularly in time according to an observational policy. Without treatment, the vital drops linearly $dx_t/dt = -(\alpha + \xi_t)$ where $\alpha > 0$ and $\xi_t\sim \gN(0, \sigma)$ is a noise term drawn at each time unit. Upon receiving treatment, the vital rises by an amount inversely proportional to the number of treatments, $1\leq k \leq m$, applied up until that time, where $m$ is the maximal number of treatments that a patient can receive. That is, the efficacy of treatment reduces with repeated applications. We also inject small noise terms into the treatment dosage that a patient receives, which further affect the vital and add randomness to the problem. \Cref{fig:simulated_trajectory} shows an example simulated patient trajectory. \textbf{(ii) Tumor growth:} We use the experimental setting from \citet{Bica2020Estimating}, which other works use to study irregular sampling \citep{seedat2022continuous, vanderschueren2023accounting}. As this is a commonly used simulator, we defer its details to \cref{app:experiments_comments} and focus on the type of irregular sampling and policies we use. The simulator works in discrete time $t\in{[T]}$, and irregular sampling is induced by the features being unobserved at certain times. Namely, the covariate $x_t\in{\mathbb{R}}_+$ which represents tumor volume is observed with probability $\sigma((\bar{x}_{t-d:t-1} / d_{\mathrm{max}}) - 1.5)$, where $\bar{x}_{t-d:t-1}$ is the average tumor volume over the last $d$ timesteps, and $d_{\mathrm{max}}$ is the maximum considered volume.


\begin{figure} 
\begin{subfigure}[c]{0.45\textwidth}
\label{fig:simulated_trajectory}
\input{Figures/patient_trajectory}
\end{subfigure}
\begin{subfigure}[c]{0.55\textwidth}
\centering
\resizebox{\textwidth}{!}{
\begin{tabular}{c ccc }
\toprule
 & ERM / MC & FQE & EDQ \\
\cmidrule(lr){2-4}
 & \multicolumn{3}{c}{$\lambda_{\mathrm{int}} = 0.1$} \\
${\color{blue}\lambda_{\obs} = 0.1} $  & $0.20 \pm 0.007$ & $0.21 \pm 0.02 $ & $0.20\pm 0.005$   \\
${\color{red}\lambda_{\obs} = 0.5}$ & $0.38\pm0.011$ & $0.23\pm 0.04$ & $\mathbf{0.20 \pm 0.006}$ \\
\midrule 
 & \multicolumn{3}{c}{$\lambda_{\mathrm{int}} = 0.5$} \\
${\color{blue}\lambda_{\obs} = 0.5}$ & $0.11 \pm 0.004$ & $0.197 \pm 0.013$ & $0.10 \pm 0.003$ \\
${\color{red}\lambda_{\obs} = 0.1}$ & $0.28 \pm 0.004$ & $0.31 \pm 0.01$ & $\mathbf{0.11 \pm 0.006}$ \\
\bottomrule
\end{tabular}
}
\end{subfigure}
\caption{\textbf{Left.} An example of a trajectory from our simulation. The blue curve denotes the value of the vital $x_t$ and red dots mark treatment times. \textbf{Right.} Normalized RMSE under the different simulation settings. The mean is taken over all points in the history of patients in the test data. Rows colored blue have $\lambda_{\obs}=\lambda_{\mathrm{int}}$, and we expect all methods to perform well since train and test distributions match. Red rows are those where the effect of an intervention needs to be estimated.}
\label{tab:simulation_results}
\end{figure}

\textbf{Outcomes and policies.} For \textbf{(i) time-to-failure}, our outcome of interest is failure time $y\in{\reals_+}$, where a patient dies if the vital drops to a value of $0$. \footnote{Note that the vital changes outside measurement times, hence death time does not generally coincide with vital measurement times.} We focus on effect estimation for interventions on a rate parameter $\lambda^a$ that controls the timing of treatment. At each time $t$ where the observed vital crosses a threshold, i.e. $x_t < r$ for some predetermined $r\in{\sR_+}$, a random time is drawn from an exponential distribution $\delta\sim\exp(\lambda^a)$ and treatment is applied at $t+\delta$. The threshold $r$ and the dosage of treatment given are also part of the policy $\pi$, yet to focus on the effects of timing we do not intervene on them in this experiment. At each experiment we observe $m$ patients treated under a policy with $\lambda^a = \lambda_{\obs}$ and aim to reason about the expected failure times under the interventional $\lambda_{\mathrm{int}}$. In \textbf{(ii) tumor-growth}, the goal is to predict $x_T$, where $T=20$, under a policy that at time $t$ assigns treatment $a_{t}$ from four possible options: no-treatment, radiotherapy, chemotherapy, or combined therapy. Therefore, the estimation here is both on ``when" and ``what" to do. Policies are determined by two parameters $(\gamma, \beta)$ and assign each type of treatment with probability $\sigma(\gamma( x_{\mathrm{last}} - \beta) + t-t_{\mathrm{last}})$. Here, $x_{\mathrm{last}}$ is the last observed volume and $\beta$ is an intercept controlling how often treatments are applied, while $\gamma$ controls the dependence of treatment assignment on tumor volume. Finally $t_{\mathrm{last}}$ is the time of the last treatment, and the term $t-t_{\mathrm{last}}$ induces a lag between consecutive treatments.

\begin{figure} 
\begin{subfigure}[c]{0.55\textwidth}
\label{fig:results2}
\resizebox{\textwidth}{!}{
\begin{tabular}{c ccc}
\toprule
 & ERM / MC & FQE & EDQ \\
 \cmidrule(lr){2-4}
 & \multicolumn{3}{c}{$(\gamma, \beta)_{\mathrm{int}} = (6, 0.75)$} \\
 ${\color{blue}\left( \gamma, \beta \right)_{\mathrm{obs}} = (6, 0.75) }$  & $\mathbf{0.034 \pm 0.001}$ & $0.048 \pm 0.001$ & $0.037 \pm 0.001$   \\
${\color{red}\left( \gamma, \beta \right)_{\mathrm{obs}} = (10, 0.5)} $  & $0.07 \pm 0.004$ & $0.080 \pm 0.013$ & $\mathbf{0.052 \pm 0.006}$   \\
\bottomrule
\end{tabular}
}
\end{subfigure}
\begin{subfigure}[c]{0.45\textwidth}
\centering
\resizebox{\textwidth}{!}{
\begin{tabular}{c ccc}
\toprule
 & ERM / MC & FQE & EDQ \\
 \cmidrule(lr){2-4}
 & \multicolumn{3}{c}{$\lambda_{\mathrm{int}} = 0.2$} \\
${\color{blue}\lambda_{\obs} = 0.2} $  & $\mathbf{0.17 \pm 0.01}$ & $0.18 \pm 0.004$ & $0.178 \pm 0.01$   \\
${\color{red}\lambda_{\obs} = 2} $ & $0.28\pm0.01$ & $0.20\pm 0.03$ & $\mathbf{0.178\pm 0.01}$ \\
\midrule
 & \multicolumn{3}{c}{$\lambda_{\mathrm{int}} = 2$} \\
${\color{blue}\lambda_{\obs} = 2}$ & $0.22 \pm 0.01$ & $\mathbf{0.197 \pm 0.013}$ & $0.22 \pm 0.004$ \\
${\color{red}\lambda_{\obs} = 0.2}$ & $0.32\pm 0.02$ & $0.31 \pm 0.01$ & $\mathbf{0.22 \pm 0.007}$ \\
\bottomrule
\end{tabular}
}
\end{subfigure}
\caption{\textbf{Left.} Normalized RMSE on the \textbf{tumor-growth} simulation. All methods are affected by distribution shift. \ours{} is the most robust out of the methods considered. \textbf{Right.} Normalized RMSE for \textbf{time-to-failure} simulation on short trajectories. 
}
\label{tab:simulation_results2}
\end{figure}
\textbf{Experiments.} We perform two sets of experiments for 
\textbf{time-to-failure}.
In the first set, trajectory lengths range in $10-100$,
and there are $5$ treaments.
In the second set (results in \Cref{tab:simulation_results2}, right), we change the parameters of the problem by taking a high slope $\alpha$ and capping
at one treatment.
This creates short trajectories of length between $3$ and $10$. To evaluate the performance of the estimator, we sample trajectories $(\gH_i, y_i) \sim P_{\lambda_{\mathrm{int}}}$ under the target policy and treat every $(\gH_{i, t_i}, y_i)$ as a labeled data point. We then evaluate normalized RMSE between $f_{\vtheta}(\gH_{i, t_i})$ and the true labels $y_i$. For \textbf{tumor growth}, we evaluate a policy that increases the likelihood of treatments (i.e increases $\beta$) and reduces $\gamma$, the correlation to the observed volume. Error is also calculated with normalized RMSE.

\textbf{Results.} The tables in \cref{tab:simulation_results} and \cref{tab:simulation_results2} present results. They show that for \textbf{time-to-failure}, \ours{} solves the estimation problem both when $\lambda_{\obs}=\lambda_{\mathrm{int}}$ (blue rows, no intervention performed), and when $\lambda_{\obs} \neq \lambda_{\mathrm{int}}$ (red rows). This is evident by comparing to ERM under the setting where $\lambda_{\obs}=\lambda_{\mathrm{int}}$, where ERM should be nearly optimal. \footnote{this is up to numerical optimization issues, as we see FQE can outperform it in certain cases} ERM takes a significant performance drop when $\lambda_{\obs} \neq \lambda_{\mathrm{int}}$, as expected. For FQE, while in the first set of experiments 
(\cref{tab:simulation_results}), discretization should not result in significant information loss, it does create a more difficult optimization problem.
This is because the updates to $Q(\gH_t)$ need to propagate backwards and most updates get noisy gradient signals by fitting to the $Q$ value of a trajectory sampled one time step ahead, $Q(\tgH_{t+1})$. This challenge for FQE is most evident in \Cref{tab:simulation_results}, where $\lambda_{\mathrm{int}}=0.5$ and FQE incurs a significant loss both when $\lambda_{\obs}=\lambda_{\mathrm{int}}$ and when $\lambda_{\obs} \neq \lambda_{\mathrm{int}}$. The results in \Cref{tab:simulation_results2} (right) demonstrate 
the potential effects of information loss due to discretization. Here, since trajectories are short, the optimization problem of losses propagating 
is likely less pronounced. However, we see that there is still a significant drop when $\lambda_{\mathrm{int}}=2$ and $\lambda_{\obs}=0.2$
(approximating a high rate of treatment when data was sampled under low rates). Taken together, these two experiments demonstrate two possible drawbacks of discretization. For \textbf{tumor-growth}, \ours{} still outperforms the alternatives but suffers a 
decrease in performance due to distribution shift between observational and interventional distributions.

\section{Limitations and Future Work}
We have developed a method for off-policy evaluation with irregular treatment and observation times, which facilitates interventions on treatment intensities. We connected the setting to identifiability results from causal inference 
to highlight the conditions under which the estimates are 
meaningful,
and proved the estimator's correctness. 
\ours{} is a ``direct" method based on regression and, as demonstrated in experiments, is readily applicable to high-capacity sequence modeling architectures. To the best of our knowledge, it is the first available solution to this estimation problem that is applied with such architectures. Several limitations motivate exciting future research. Empirically, we plan to apply the method to large real-world datasets and additional simulators \citep{oberst2019counterfactual, namkoong2020off}.
The method does not handle censoring, which is required for
reliable application in
survival analysis and real trial data. Further developements include 
policy optimization in the setting we studied here and deriving bounds on errors due to unobserved confounding.

\subsubsection*{Acknowledgments}
We thank Stefan Groha for helpful discussion in early stages of the project. This work was supported by National
Science Foundation Award 1922658, NIH/NHLBI Award
R01HL148248, NSF CAREER Award 2145542, NSF Award 2404476, ONR N00014-23-1-2634, IITP with a grant from ROK-MSIT in connection with the Global AI Frontier Lab International Collaborative Research, Apple, and Optum.

%% file: Figures/patient_trajectory.tex
\begin{tikzpicture}[scale=0.6]
    \begin{axis}[
        xlabel={$t$},
        ylabel={$x$},
        xmin=0, xmax=70,
        ymin=0, ymax=7,
        legend pos=north east,
        legend cell align={left},
        legend style={draw=none},
        xlabel style={
            font=\large,
        },
        ylabel style={
            font=\large,    
            yshift=-12pt     
        },
    ]
    ]
    \addplot[
        thick,
        blue,
    ] coordinates {
        (0.0, 6.792586177866876) (1.0, 6.502659536958309) (2.0, 6.37376509458887) (3.0, 6.250331975244174) (4.0, 6.065064860378735) (5.0, 5.78368126425061) (6.0, 5.565059308150195) (7.0, 5.393195429283178) (8.0, 5.275660849324314) (9.0, 4.938863119445168) (10.0, 4.559648231143505) (11.0, 4.364391962442589) (12.0, 3.9803581674069575) (13.0, 3.8264928510001415) (14.0, 3.5831523522871502) (15.0, 3.3188420034371884) (16.0, 2.9248135367930903) (17.0, 2.7464384516777454) (18.0, 2.3891216390848498) (19.0, 2.2133922414326195) (20.0, 2.048603602506043) (21.0, 1.927355435911696) (22.0, 1.8095560618576085) (23.0, 6.632160785101375) (24.0, 6.419710909909608) (25.0, 6.159914547325677) (26.0, 5.861953069939286) (27.0, 5.64075239641299) (28.0, 5.486978785324833) (29.0, 5.321406310880468) (30.0, 5.075924310570866) (31.0, 4.804330748893358) (32.0, 4.426410918929223) (33.0, 4.287834895788054) (34.0, 4.1019934554112005) (35.0, 3.8572845601898704) (36.0, 3.5296702947673886) (37.0, 3.3198717727274225) (38.0, 3.085481707883489) (39.0, 2.6890650710558313) (40.0, 2.5855910360525827) (41.0, 2.4602235309485327) (42.0, 2.2954225294986315) (43.0, 1.994505083576481) (44.0, 1.783833775114753) (45.0, 3.7260507566424375) (46.0, 3.3514958547964464) (47.0, 3.1905010481631053) (48.0, 2.929600277223791) (49.0, 2.8091077203591297) (50.0, 2.6173588347811183) (51.0, 2.218641352922984) (52.0, 1.991140909119061) (53.0, 1.8236859682858861) (54.0, 1.699433326779056) (55.0, 1.356997163696874) (56.0, 2.8722599307467753) (57.0, 2.7210017954284145) (58.0, 2.323352934056665) (59.0, 2.0400042609478555) (60.0, 1.8198484024621524) (61.0, 1.700746587759908) (62.0, 1.5554585295232817) (63.0, 1.2952637329447718) (64.0, 1.1066846467071922) (65.0, 2.1256355416156048) (66.0, 1.7319418567414846) (67.0, 1.376213187713946) (68.0, 1.0418293850015998) (69.0, 0.7313082116975171) (70.0, 0.44363243489856324) (71.0, 0.3386296829261567) (72.0, 1.1251918580118727) (73.0, 1.0009747237352553) (74.0, 0.8267831068676821) (75.0, 0.5731705038123689) (76.0, 0.4465039445751089) (77.0, 0.2225337438742182)
    };
    \addlegendentry{Vital $x_t$}

    \addplot[
        only marks,
        red,
        mark=*,
        mark size=2.5pt
    ] coordinates {
        (22.85286175758743, 0) (44.33163603286685, 0) (55.09203742174845, 0) (64.9717165996269, 0) ( 71.02345131600424, 0)
    };
    \addlegendentry{Treatment}

    \end{axis}
\end{tikzpicture}

%% file: sections/A_proofs.tex
\section{Proofs}
We begin with some notation and additional definitions, in \cref{sec:tower_expectation} we prove the consistency result for our method, and in \cref{sec:tower_expectation_discrete} we give its discrete-time version. To avoid cluttered notation and longer proof, we will give the proof of \cref{thm:expectation_identity} for unmarked processes. Adding a distribution of marks is a trivial extension that does not alter the main steps of the derivation.

\subsection{Notation and Definitions} \label{sec:app_notation}
For a multivariate point process we use the following notations:
\begin{itemize}
    \item $\lambda_{\bullet}(\cdot)$ is the sum $\sum_{e}{\lambda^{e}(\cdot)}$, in our case this will include the components $e\in\{a, x, y\}$.
    \item For any $s > t$ and any distribution or intensity, e.g. $\lambda$, we will use the conditioning $\lambda(\cdot \vert \gH_s=\gH_t)$ to denote the event where jumps until time $t+\delta$ are those that appear in $\gH_t$. That is, no events occur in the interval $(t, s]$.
    \item $\gH_t \cup \{(t+\delta, e)\}$ is the event in which jumps until time $t+\delta$ are those that appear in $\gH_t$, and the next jump after that happens at time $t+\delta$ and is of type $e$ (i.e. $N^e(t+\delta) = N^e(t) + 1$ and $N^e(t+s)= N^e(t)$ for $s\in(0,\delta)$).
    \item Given a trajectory $\tgH$ and time $t$, we define $\delta^{e}(t) = \min\{s-t : s > t, (s,\cdot)\in{\tgH^e}\}$ as the time gap from time $t$ to the first jump of process $N_e$ in trajectory $\gH$ after $t$, that is for $e\in{\{x,y,a,a_{\obs}\}}$. \\
    \emph{Note:} This is a slight abuse of notation from the main paper, where we defined $\delta_{\tgH}(t) = \min\{u-t : u > t, (u,\cdot)\in{\tgH}^{a,a_{\obs}}\}$: dependence on $\tgH$ is omitted and will be included whenever the trajectory is not clear from context, and we add a superscript (e.g. $\delta^{a,a_{\obs}}$) to specify the type of event we look for.
    \item Since the processes $N^x,N^y$ play a similar role throughout the derivation, as the parts of the process whose intensities are invariant under the intervention, we will shorten notation to
    \begin{align*}
        &\lambda^{v}(t+\delta \vert \gH_{t+\delta})\E_{P}\left[ Y \vert \gH_t \cup \{ (t+\delta, v) \} \right] := \\ &\quad \lambda^x(t+\delta \vert \gH_{t+\delta})\E_{P}\left[ Y \vert \gH_t \cup \{ (t+\delta, x) \} \right] + \lambda^y(t+\delta \vert \gH_{t+\delta})\E_{P}\left[ Y \vert \gH_t \cup \{ (t+\delta, y) \} \right], 
    \end{align*}
    and $\delta^v := \delta_x \wedge \delta_y$.
    
\end{itemize} 
We assume that all processes have well-defined densities and intensity functions, and that $P$ is absolutely continuous w.r.t. $P_\obs$, $P \ll P_\obs$. This means that the conditional expectations taken w.r.t. $P$, which we use in our derivation, are well defined.
We also adopt the convention where $N^e(t)$ is almost surely finite for any $t\in{[0,T]}$ and $e\in{\{ a,x,y \}}$ 
\citep{andersen2012statistical}. This means that the number of events in the interval $[0,T]$ is countable. We also use the notation $\mathbf{1}[\cdot]$ for the indicator function that returns $1$ if the condition inside it is satisfied and $0$ otherwise.


\subsection{Proof of Formal Results} \label{sec:tower_expectation}
Below we prove \cref{thm:expectation_identity}
where the result, \cref{eq:expectation_identity}, implies that performing dynamic programming using the $Q$-function from the earliest disagreement time between observed data, and the data sampled from the target distribution, results in a correct estimator.
We derive that equation from the lemma below, which is similar to a tower property of conditional expectations with respect to the first jump that occurs in any component of the process.

\begin{lemma} \label{lem:one_stop_lemma}
Let $P, P_\obs$ be multivariate marked decision point processes, $\tilde{P}$ the corresponding augmented process, $t\in{[0, T)}$,
and $\gH_t$ a history of events that is measurable w.r.t $P$.
It holds that
\begin{align} \label{eq:alg_expect_general}
\E_{P}[Y\vert \gH_t] = \E_{\tgH\sim \tPr(\cdot  \vert \gH_t)}\Big[ & \mathbf{1}\big[\delta^a(t) < \delta^v(t) \wedge \delta^{a_{\obs}}(t)\big]\E_P\left[ Y \vert \gH_t \cup (t+\delta^a(t), a) \right] + \nonumber \\
 & \mathbf{1}\big[\delta^v(t) < \delta^a(t)\wedge \delta^{a_{\obs}}(t)\big]\E_P\left[ Y \vert \gH_t \cup (t+\delta^v(t), v) \right] + \nonumber \\ 
 & \mathbf{1}\big[\delta^{a_{\obs}}(t) < \delta^a(t) \wedge \delta^v(t) \big]\E_P\left[ Y \vert \gH_{t+\delta^{a_{\obs}}(t)} = \gH_t \right] + \nonumber \\
 & \mathbf{1}\big[\delta^{a_{\obs}}(t)\wedge\delta^a(t)\wedge \delta^v(t) > T-t\big]\E_P\left[ Y \vert \gH_{T} = \gH_t \right]\Big]\Big.
\end{align}
\end{lemma}
\begin{proof}

Note that all the conditional expectations in the above expression exist since $P \ll P_\obs$. Denoting the next jump time with a variable $T_{\mathrm{next}}$ and its type by $E_{\mathrm{next}}$, when conditioning on some history $\gH_{s}$, the law of total probability suggests that
$P(Y \vert \gH_t) = \int \sum_{e\in{\{\rma, \rmx, \rmy\}}} P(Y \vert \gH_t, T_{\mathrm{next}}=t+\delta, E_{\mathrm{next}=e})P(T_{\mathrm{next}}=t+\delta, E_{\mathrm{next}=e} \vert \gH_t)\rmd t$. In point processes, likelihoods of the form 
$P(T_{\mathrm{next}}=t+\delta, E_{\mathrm{next}}=e \vert \gH_t)$ are given by $\exp\{ -\int_{t}^{t+\delta}\lambda_{\bullet} (s \vert \gH_s=\gH_t)\}\cdot \lambda^{e}(t+\delta \vert \gH_{t+\delta}=\gH_t)$. Expanding $\E_P[Y \vert \gH_t]$ with the law of total probability and these likelihoods, while accounting for the option that no jump occurs in $(t,T]$, we obtain the following expression.
\begin{align}
    \E_{P}\left[ Y \vert \gH_t \right] = \exp&\left\{ -\int_{t}^{T}\lambda_{\bullet}(s \vert \gH_s=\gH_t) \right\}\E_{P}\left[ Y \vert \gH_T=\gH_{t} \right] + \nonumber \\
    \int_{0}^{T-t} \exp&\left\{ -\int_{t}^{t+\delta} \lambda_{\bullet}(s \vert \gH_{s} = \gH_t) \rmd s \right\} \nonumber \\
    &\Big ( \lambda^a(t+\delta \vert \gH_{t+\delta} = \gH_t) \E_{P}\left[ Y \vert \gH_{t} \cup \{(t+\delta, a)\} \right] + \nonumber \\
    & \:\:\: \lambda^x(t+\delta \vert \gH_{t+\delta} = \gH_t) \E_{P}\left[ Y \vert \gH_{t} \cup \{(t+\delta, x)\} \right] + \nonumber \\
    & \:\:\: \lambda^y(t+\delta \vert \gH_{t+\delta} = \gH_t) \E_{P}\left[ Y \vert \gH_{t} \cup \{(t+\delta, y)\} \right] \Big) \rmd \delta.
    \label{eq:target_estimand}
    \end{align}

Next we write down each item in \cref{eq:alg_expect_general},
\begin{align} \label{eq:item_one}
    &\E_{\tgH\sim \tPr(\cdot \vert \gH_t)}\Big[
    \mathbf{1}\big[ \delta^a(t) < \delta^v(t)\wedge \delta^{a_{\obs}}(t)\big]\cdot \E\left[ Y \vert \gH_t \cup (t+\delta^a(t), a) \right] \Big] = \nonumber \\
    &\quad \int_{0}^{T-t} \lambda^a(t+\delta \vert \gH_{t+\delta} = \gH_t)\exp\{-\int_{t}^{t+\delta}\lambda^a(s \vert \gH_{s} = \gH_t) \rmd s\} \nonumber \\ &\qquad \qquad \exp\{-\int_{t}^{t+\delta} \lambda_{\obs, \bullet}(s \vert \gH_s = \gH_t)\rmd s\} \E\left[ Y \vert \gH_t \cup (t+\delta, a) \right] \rmd \delta = \nonumber \\
    &\quad \int_{0}^{T-t} \lambda^a(t+\delta \vert \gH_{t+\delta} = \gH_t)\exp\{-\int_{t}^{t+\delta}\lambda_{\bullet}(s \vert \gH_{s} = \gH_t) \rmd s\} \nonumber \\
    &\qquad \qquad\exp\{-\int_{t}^{t+\delta} \lambda_{\obs}^a(s \vert \gH_s = \gH_t)\rmd s\} \E\left[ Y \vert \gH_t \cup (t+\delta, a) \right] \rmd \delta = \nonumber \\
    &\quad \int_{0}^{T-t} \lambda^a(t+\delta \vert \gH_{t+\delta} = \gH_t)\exp\{-\int_{t}^{t+\delta}\lambda_{\bullet}(s \vert \gH_{s} = \gH_t) \rmd s\}\E\left[ Y \vert \gH_t \cup (t+\delta, a) \right] \nonumber \\
    & \qquad \qquad \cdot (1 -1 + \exp\{-\int_{t}^{t+\delta} \lambda_{\obs}^a(s \vert \gH_s = \gH_t)\rmd s\}) \rmd \delta.
\end{align}
The first equality simply expands the expectation as an integration over all possible stopping times for $N_a$ (according to the definition of $\tPr$, see \cref{def:future_dist}). The second equality holds since the intensities $\lambda_{\obs}^{x},\lambda_{\obs}^{y}$ are equal to $\lambda_{x},\lambda_{y}$ respectively. Then finally we simply add and subtract $1$ from the last item.
Similarly, for the second item in \cref{eq:alg_expect_general}
\begin{align} \label{eq:item_two}
&\E_{\tgH\sim \tPr(\cdot \vert \gH_t)}\Big[
{\mathbf{1}\big[\delta^v(t) < \delta^a(t)\wedge \delta^{a_{\obs}}(t)\big]\cdot \E\left[ Y \vert \gH_t \cup (t+\delta^v(t), v) \right] \Big]} = \nonumber \\
    &\quad \int_{0}^{T-t} \lambda^v(t+\delta \vert \gH_{t+\delta} = \gH_t)\exp\{-\int_{t}^{t+\delta}\lambda_{\bullet}(s \vert \gH_{s} = \gH_t) \rmd s\}\E\left[ Y \vert \gH_t \cup (t+\delta, v) \right] \nonumber \\
    &\qquad \qquad \cdot (1 -1 + \exp\{-\int_{t}^{t+\delta} \lambda^{a}_{\obs}(s \vert \gH_s = \gH_t)\rmd s\}) \rmd \delta.
\end{align}
The last item in \cref{eq:alg_expect_general} is
\begin{align} \label{eq:item_four}
&\E_{\tgH\sim \tPr(\cdot \vert \gH_t)}\Big[\mathbf{1}\big[\delta^{a_{\obs}}(t)\wedge\delta^a(t)\wedge \delta^v(t) > T-t\big]\cdot\E_P\left[ Y \vert \gH_{T} = \gH_t \right]\Big] =  \nonumber \\ 
&\left(1- 1 +\exp\left\{ -\int_{t}^{T}\lambda_{\obs}^{a}(s \vert \gH_s=\gH_t) \right\}\right)\exp\left\{ -\int_{t}^{T}\lambda_{\bullet}(s \vert \gH_s=\gH_t) \rmd s \right\}\E_{P}\left[ Y \vert \gH_T = \gH_{t} \right]
\end{align}
Adding up \cref{eq:item_one}, \cref{eq:item_two} and \cref{eq:item_four} while pulling out all the items multiplied by $1$ to the last brackets in the bottom expression, we get,
\begin{align*}
    &\E_{\tgH\sim \tPr(\cdot \vert \gH_t)}\Big[  \mathbf{1}\big[\delta^a(t) < \delta^v(t)\wedge \delta^{a_{\obs}}(t)\big]\cdot\E_P\left[ Y \vert \gH_t \cup (t+\delta^a(t), a) \right] + \nonumber \\
     &\:\quad\qquad\qquad \mathbf{1}\big[\delta^v(t) < \delta^a(t)\wedge \delta^{a_{\obs}}(t)\big]\cdot\E_P\left[ Y \vert \gH_t \cup (t+\delta^v(t), v) \right] + \nonumber \\ 
     &\:\quad\qquad\qquad \mathbf{1}\big[\delta^{a_{\obs}}(t)\wedge\delta^a(t)\wedge \delta^v(t) > T-t\big]\cdot\E_P\left[ Y \vert \gH_{T} = \gH_t \right]\Big] = \nonumber \\
     &\quad \Bigg( \int_{0}^{T-t} \lambda^a(t+\delta \vert \gH_{t+\delta} = \gH_t)\exp\{-\int_{t}^{t+\delta}\lambda_{\bullet}(s \vert \gH_{s} = \gH_t) \rmd s\}\E\left[ Y \vert \gH_t \cup (t+\delta, a) \right] \nonumber \\
    &\qquad \qquad \cdot (-1 + \exp\{-\int_{t}^{t+\delta} \lambda^{a}_{\obs}(s \vert \gH_s = \gH_t)\rmd s\}) \rmd \delta\Bigg) \\
    & + \Bigg( \int_{0}^{T-t} \lambda^v(t+\delta \vert \gH_{t+\delta} = \gH_t)\exp\{-\int_{t}^{t+\delta}\lambda_{\bullet}(s \vert \gH_{s} = \gH_t) \rmd s\}\E\left[ Y \vert \gH_t \cup (t+\delta, v) \right] \nonumber \\
    &\qquad \qquad \cdot (-1 + \exp\{-\int_{t}^{t+\delta} \lambda^{a}_{\obs}(s \vert \gH_s = \gH_t)\rmd s\}) \rmd \delta \Bigg)\\
    & + \Bigg( \exp\left\{ -\int_{t}^{T}\lambda_{\bullet}(s \vert \gH_s=\gH_t) \rmd s \right\}\E_{P}\left[ Y \vert \gH_T = \gH_{t} \right] \\
    & \qquad \qquad \cdot \Big(- 1 +\exp\left\{ -\int_{t}^{T}\lambda_{\obs}^{a}(s \vert \gH_s=\gH_t) \right\}\Big)\Bigg) \\
    & + \Bigg( \int_{0}^{T-t} \lambda^a(t+\delta \vert \gH_{t+\delta} = \gH_t)\exp\{-\int_{t}^{t+\delta}\lambda_{\bullet}(s \vert \gH_{s} = \gH_t) \rmd s\}\E\left[ Y \vert \gH_t \cup (t+\delta, a) \right] \\
    & \qquad + \int_{0}^{T-t} \lambda^v(t+\delta \vert \gH_{t+\delta} = \gH_t)\exp\{-\int_{t}^{t+\delta}\lambda_{\bullet}(s \vert \gH_{s} = \gH_t) \rmd s\}\E\left[ Y \vert \gH_t \cup (t+\delta, v) \right] \\
    & \qquad + \exp\left\{ -\int_{t}^{T}\lambda_{\bullet}(s \vert \gH_s=\gH_t) \rmd s \right\}\E_{P}\left[ Y \vert \gH_T = \gH_{t} \right]\Bigg). \\
\end{align*}
The items in the last brackets equal the right-hand-side of \cref{eq:target_estimand}. Plugging this in we rewrite,
\begin{align}
    \E_{\tgH\sim \tPr(\cdot \vert \gH_t)}\Big[ & \mathbf{1}\big[\delta^a(t) < \delta^v(t)\wedge \delta^{a_{\obs}}(t)\big]\cdot\E_P\left[ Y \vert \gH_t \cup (t+\delta^a(t), a) \right] + \nonumber \\
     & \mathbf{1}\big[\delta^v(t) < \delta^a(t)\wedge \delta^{a_{\obs}}(t)\big]\cdot\E_P\left[ Y \vert \gH_t \cup (t+\delta^v(t), v) \right] + \nonumber \\ 
     & \mathbf{1}\big[\delta^{a_{\obs}}(t)\wedge\delta^a(t)\wedge \delta^v(t) > T-t\big]\cdot\E_P\left[ Y \vert \gH_{T} = \gH_t \right]\Big] = \nonumber \\
     \E_{P}\left[ Y \vert \gH_t \right] + \qquad\qquad\qquad& \nonumber \\ - \Bigg(1-\exp\Bigg\{ -\int_{t}^{T}\lambda^a_{\obs}&(s \vert \gH_s=\gH_t) \rmd s \Bigg\}\Bigg) \exp\left\{ -\int_{t}^{T}\lambda_{\bullet}(s \vert \gH_s=\gH_t) \rmd s \right\} \E_{P}[Y \vert \gH_T=\gH_t] \nonumber \\
     - \int_{0}^{T-t}\Bigg[\Big(\lambda^v(t+\delta \vert \gH_{t+\delta} &= \gH_t)\E_{P}[Y \vert \gH_t \cup (t+\delta, x)] \nonumber \\ + \lambda^a(t+\delta \vert \gH_{t+\delta} &= \gH_t)\E_{P}[Y \vert \gH_t \cup (t+\delta, a)]\Big) \nonumber \\
    \cdot\exp\{-\int_{t}^{t+\delta}\lambda_{\bullet}(s \vert \gH_{s} &= \gH_t) \rmd s\}
    \cdot \left( 1 - \exp\{-\int_{t}^{t+\delta} \lambda_{\obs}^a(s \vert \gH_s = \gH_t)\rmd s\} \right)\Bigg]\rmd \delta.    \label{eq:expanding_items_1_2}
\end{align}
Note that we have,
\begin{align}
\label{eq:prob_jump_in_delta}
    &1 - \exp\{-\int_{t}^{t+\delta} \lambda_{\obs}^a(s \vert \gH_s =  \gH_t)\rmd s\} = \nonumber\\
    &\qquad \int_{t}^{t+\delta}{\lambda^a_{\obs}(s \vert \gH_s=\gH_t)\exp\{-\int_{t}^{s}\lambda_{\obs}^a(s \vert\gH_s=\gH_t) \rmd u}\} \rmd s,
\end{align}
because the left-hand-side is $1$ minus the probability that $N_a^{\obs}$ does not jump in the interval $(t, t+\delta]$, and the integration on the right hand side is the probability that the process jumps at least once (where the first jump is at time $s$).

Next, we write the third item of \cref{eq:alg_expect_general} to see that it cancels the residual above in \cref{eq:expanding_items_1_2}.
\begin{align}
&\E_{\tgH\sim \tPr(\cdot \vert \gH_t)}\Big[{\mathbf{1}\big[\delta^{a_{\obs}}(t) < \delta^a(t)\wedge \delta^v(t)\big]\cdot\E\left[ Y \vert \gH_{t+\delta^{a_{\obs}}} = \gH_t \right]}\Big] = \nonumber \\
& \quad \int_{0}^{T-t} \lambda_{\obs}^a(t+\delta \vert \gH_{t+\delta} = \gH_t)\exp\{-\int_{t}^{t+\delta}\lambda_{\obs}^a(s \vert \gH_{s} = \gH_t) \rmd s\} \nonumber \\
& \qquad \qquad {\color{blue}\exp\{-\int_{t}^{t+\delta} \lambda_{\bullet}(s \vert \gH_s = \gH_t)\rmd s\} \E_P\left[ Y \vert \gH_{t+\delta}= \gH_t \right]}\rmd \delta.
\label{eq:third_item}
\end{align}
We expand $\E_P\left[ Y \vert \gH_{t+\delta}= \gH_t \right]$ again by towering expectations w.r.t to the first jump after $t+\delta$,
\begin{align*}
\E_P[ Y \vert \gH_{t+\delta}&= \gH_t ] = \int_{t+\delta}^{T} \Big(\lambda^a(s' \vert \gH_s= \gH_t)\E_{P}\left[ Y \vert \gH_{s'} = \gH_t \cup (s', a) \right] + \nonumber \\
&\lambda^v(s' \vert \gH_{s'}= \gH_t)\E_{P}\left[ Y \vert \gH_{s'} = \gH_t \cup (s', v) \right]\Big)\exp\{-\int_{t+\delta}^{s'} \lambda_{\bullet}(u \vert \gH_u= \gH_t) \rmd u\}\rmd s' \\
&+\E_{P}[Y \vert \gH_T=\gH_t]\exp\{ -\int_{t+\delta}^{T} \lambda_{\bullet}(u \vert \gH_{u}=\gH_t)\rmd u\}
\end{align*}
Multiplying the left hand side by $\exp\{ - \int_{t}^{t+\delta}\lambda_{\bullet}(s \vert \gH_s=\gH_t)\rmd s\}$ we get a similar item where the integration on $\lambda_{\bullet}$ starts from $t$ instead of $t+\delta$,
\begin{align*}
\color{blue}\exp\{-\int_{t}^{t+\delta} &{\color{blue}\lambda_{\bullet}(s \vert \gH_s = \gH_t)\rmd s\} \E_P\left[ Y \vert \gH_{t+\delta}= \gH_t \right]} = \\
{\color{gray}\int_{t+\delta}^{T} \Big(}&{\color{gray}\lambda^a(s' \vert \gH_{s'}= \gH_t)\E_{P}\left[ Y \vert \gH_{s'} = \gH_t \cup (s', a) \right] +} \nonumber \\
&{\color{gray}\lambda^v(s' \vert \gH_{s'}= \gH_t)\E_{P}\left[ Y \vert \gH_{s'} = \gH_t \cup (s', v) \right]\Big)\exp\{-\int_{t}^{s'} \lambda_{\bullet}(u \vert \gH_u= \gH_t) \rmd u\}\rmd s'} \\
&+\E_{P}[Y \vert \gH_T=\gH_t]\exp\{ -\int_{t}^{T} \lambda_{\bullet}(u \vert \gH_u=\gH_t)\rmd u\}
\end{align*}
Let us denote the gray item by ${\color{gray}(*)}$, and multiply by the observed part in \cref{eq:third_item}. In the first equality we will pull the integration on $s'$ outside, then we will change the order of integration, change variables by a constant shift ($\delta'=s'-t, \tilde{s}=\delta+t$), and push one integration back inside.
\begin{align*}
    \int_{0}^{T-t}&\lambda_{\obs}^a(t+\delta \vert \gH_{t+\delta} = \gH_t)\exp\{-\int_{t}^{t+\delta}\lambda_{\obs}^a(s \vert \gH_{s} = \gH_t) \rmd s\}\cdot {\color{gray}{(*)}} = \\
    \int_{0}^{T-t}&\lambda_{\obs}^a(t+\delta \vert \gH_{t+\delta} = \gH_t)\exp\{-\int_{t}^{t+\delta}\lambda_{\obs}^a(s \vert \gH_{s} = \gH_t) \rmd s\}\cdot \\
    \Bigg[\int_{t+\delta}^{T} \Big(&\lambda^a(s' \vert \gH_{s'}= \gH_t)\E_{P}\left[ Y \vert \gH_{s'} = \gH_t \cup (s', a) \right] + \nonumber \\
\lambda^v(s' \vert &\gH_{s'}= \gH_t)\E_{P}\left[ Y \vert \gH_{s'} = \gH_t \cup (s', v) \right]\Big)\exp\{-\int_{t}^{s'} \lambda_{\bullet}(u \vert \gH_u= \gH_t) \rmd u\}\rmd s'\Bigg] \rmd \delta \\
= \int_{0}^{T-t}\int_{t+\delta}^{T}&\lambda_{\obs}^a(t+\delta \vert \gH_{t+\delta} = \gH_t)\exp\{-\int_{t}^{t+\delta}\lambda_{\obs}^a(s \vert \gH_{s} = \gH_t) \rmd s\}\cdot \\
\Big(&\lambda^a(s' \vert \gH_{s'}= \gH_t)\E_{P}\left[ Y \vert \gH_{s'} = \gH_t \cup (s', a) \right] + \nonumber \\
\lambda^v(s' \vert &\gH_{s'}= \gH_t)\E_{P}\left[ Y \vert \gH_{s'} = \gH_t \cup (s', v) \right]\Big)\exp\{-\int_{t}^{s'} \lambda_{\bullet}(u \vert \gH_u= \gH_t) \rmd u\} \rmd s' \rmd\delta \\
= \int_{t}^{T}\int_{0}^{s'-t}&\lambda_{\obs}^a(t+\delta \vert \gH_{t+\delta} = \gH_t)\exp\{-\int_{t}^{t+\delta}\lambda_{\obs}^a(s \vert \gH_{s} = \gH_t) \rmd s\}\cdot \\
\Big(&\lambda^a(s' \vert \gH_{s'}= \gH_t)\E_{P}\left[ Y \vert \gH_{s'} = \gH_t \cup (s', a) \right] + \nonumber \\
\lambda^v(s' \vert &\gH_{s'}= \gH_t)\E_{P}\left[ Y \vert \gH_{s'} = \gH_t \cup (s', v) \right]\Big)\exp\{-\int_{t}^{s'} \lambda_{\bullet}(u \vert \gH_u= \gH_t) \rmd u\} \rmd\delta \rmd s' \\
= \int_{0}^{T-t}\int_{t}^{t+\delta'}&\lambda_{\obs}^a(\tilde{s} \vert \gH_{\tilde{s}} = \gH_t)\exp\{-\int_{t}^{\tilde{s}}\lambda_{\obs}^a(s \vert \gH_{s} = \gH_t) \rmd s\}\cdot \\
\Big(&\lambda^a(t+\delta' \vert \gH_{t+\delta'}= \gH_t)\E_{P}\left[ Y \vert \gH_{t+\delta'} = \gH_t \cup (t+\delta', a) \right] + \nonumber \\
&\lambda^v(t+\delta' \vert \gH_{t+\delta'}= \gH_t)\E_{P}\left[ Y \vert \gH_{t+\delta'} = \gH_t \cup (t+\delta', v) \right]\Big) \cdot \\
&\exp\{-\int_{t}^{t+\delta'} \lambda_{\bullet}(u \vert \gH_u= \gH_t) \rmd u\} \rmd \tilde{s} \rmd\delta'
\\
= {\color{red} \int_{0}^{T-t} \Big(}&{\color{red}\lambda^a(t+\delta' \vert \gH_{t+\delta'}= \gH_t)\E_{P}\left[ Y \vert \gH_{t+\delta'} = \gH_t \cup (t+\delta', a) \right] + }\nonumber \\
&{\color{red}\lambda^v(t+\delta' \vert \gH_{t+\delta'}= \gH_t)\E_{P}\left[ Y \vert \gH_{t+\delta'} = \gH_t \cup (t+\delta', v) \right]\Big) \cdot} \\
{\color{red}\exp\{-\int_{t}^{t+\delta'}}& {\color{red}\lambda_{\bullet}(u \vert \gH_u= \gH_t) \rmd u\} \left(\int_{t}^{t+\delta'}\lambda_{\obs}^a(\tilde{s} \vert \gH_{\tilde{s}} = \gH_t)\exp\{-\int_{t}^{\tilde{s}}\lambda_{\obs}^a(s \vert \gH_{s} = \gH_t) \rmd s\} \rmd \tilde{s}\right) \rmd\delta'}
\end{align*}
Plugging everything back into \cref{eq:third_item}, we color in red the same item colored red above, where we change the name of variables back from $\delta',\tilde{s}$ to $\delta, s$ for convenience. The remaining item in the equation is obtained by collecting all the items that multiply $\E_{P}[Y \vert \gH_T = \gH_t]$ in the obtained expression.
\begin{align*}
&\E_{\tgH\sim \tPr(\cdot \vert \gH_t)}\Big[{\mathbf{1}\big[\delta^{a_{\obs}}(t) < \delta^a(t)\wedge \delta^v(t)\big]\E\left[ Y \vert \gH_{t+\delta^{a_{\obs}}} = \gH_t \right]}\Big] = \\
& {\color{red} \int_0^{T-t}\Big(\lambda^a(t+\delta \vert \gH_{t+\delta}= \gH_t)\E_{P}\left[ Y \vert \gH_{t+\delta} = \gH_t \cup (t+\delta, a) \right] + }\\
&\qquad \qquad \:\: {\color{red} \lambda^v(t+\delta \vert \gH_{t+\delta}= \gH_t)\E_{P}\left[ Y \vert \gH_{t+\delta} = \gH_t \cup (t+\delta, x) \right]\Big) } \\
&{\color{red} \exp\{-\int_{t}^{t+\delta} \lambda_{\bullet}(s \vert \gH_s= \gH_t) \rmd s\} \Bigg(\int_{t}^{t+\delta}\lambda^a_{\obs}(s \vert \gH_s = \gH_t)\exp\{-\int_{t}^{s}\lambda^a_{\obs}(u \vert\gH_u = \gH_t)\rmd u\}\rmd s \Bigg) \rmd \delta} \\
&+\left(\int_{0}^{T-t} \lambda_{\obs}^a(t+\delta \vert \gH_{t+\delta} = \gH_t)\exp\{-\int_{t}^{t+\delta}\lambda_{\obs}^a(s \vert \gH_{s} = \gH_t) \rmd s\} \rmd \delta \right)\cdot \\
&\qquad \E_{P}[Y \vert \gH_T=\gH_t]\exp\{-\int_{t}^{T}\lambda_{\bullet}(s \vert \gH_s=\gH_t)\rmd s\} = \\
& \int_0^{T-t}\Big(\lambda^a(t+\delta \vert \gH_{t+\delta}= \gH_t)\E_{P}\left[ Y \vert \gH_{t+\delta} = \gH_t \cup (t+\delta, a) \right] + \\
&\qquad \qquad \:\: \lambda^v(t+\delta \vert \gH_{t+\delta}= \gH_t)\E_{P}\left[ Y \vert \gH_{t+\delta} = \gH_t \cup (t+\delta, x) \right]\Big) \\
& \exp\{-\int_{t}^{t+\delta} \lambda_{\bullet}(s \vert \gH_s= \gH_t) \rmd s\} \Bigg(1 - \exp\{-\int_{t}^{t+\delta} \lambda_{\obs}^a(s \vert \gH_s =  \gH_t)\rmd s\} \Bigg) \rmd \delta \\
&+\left(1 - \exp\{-\int_{t}^{T} \lambda_{\obs}^a(s \vert \gH_s =  \gH_t)\rmd s\}\right)\cdot \E_{P}[Y \vert \gH_T=\gH_t]\exp\{-\int_{t}^{T}\lambda_{\bullet}(s \vert \gH_s=\gH_t)\rmd s\}.
\end{align*}
In the last equality we plugged in \cref{eq:prob_jump_in_delta}. Now it can be seen that the above expression cancels with the residual of \cref{eq:expanding_items_1_2}, which means that \cref{eq:alg_expect_general} holds as claimed.
\end{proof}
As we explain in the sequel, \cref{thm:expectation_identity} follows directly from the lemma below.
\begin{lemma} \label{lem:k_stops_lemma}
For any $t\in{[0,T)}$ and $k\in{\mathbb{N}}$, define $\delta^k(t)>0$ such that $t+\delta^k(t)$ is the time of the $k$-th event after $t$ in a trajectory $\tgH$, where $\delta^0(t) = 0$ as an edge case.\footnote{as explained in \cref{sec:app_notation}, the full notation should be $\delta^k_{\tgH}(t)$, but $\tgH$ will be clear from context.} That is, assuming $\tgH = \{(t_j, v_j)\}_{j\in{\sN}}$ then $\delta^k_{\gH}(t) := \min \{t_j - t : t_{j - k + 1} > t  \}$. Analogously, we define $\delta^{k,e}(t)$ as $\delta^k_{\tgH^e}(t)$, which is the $k$-th event of type $e$. For all $d\in{\sN_+}$ we have that
\begin{align}\label{eq:k_stops_lemma}
\E_P[Y \vert \gH_t] &= \E_{\tgH\sim \tPr(\cdot \vert \gH_t)}\bigg[ \\
&\sum_{k=1}^{d} \Big( \mathbf{1}\big[\delta^{k-1, v}(t) \leq \delta^a(t) < \delta^{k,v}(t) \wedge \delta^{a_{\obs}}(t) \wedge T-t \big] \E_{P}\big[Y \vert \gH_{t} \cup \tgH^{\setminus a_{\obs}}_{\left(t,t+\delta^{a}(t)\right]}\big] \nonumber \\
&+\mathbf{1}\big[\delta^{k-1, v}(t) \leq \delta^{a_{\obs}}(t) < \delta^{k, v}(t) \wedge \delta^{a}(t) \wedge T-t \big] \E_{P}\big[Y \vert \gH_{t} \cup \tgH^{\setminus a_{\obs}}_{\left(t,t+\delta^{a_{\obs}}(t)\right]}\big]\Big) \nonumber \\
&+\mathbf{1}\big[ T-t < \delta^{d,v}(t)\wedge \delta^{a_\obs}(t)\wedge \delta^{a}(t) \big] \E_{P}[Y \vert \gH_{t} \cup \tgH^{\setminus a_{\obs}}_{\left(t,T\right]}]\nonumber \\
&+\mathbf{1}\big[\delta^{d, v}(t) < \delta^{a_{\obs}}(t) \wedge \delta^{a}(t) \wedge T-t \big] \E_{P}\big[Y \vert \gH_{t} \cup \tgH^{\setminus a_{\obs}}_{\left(t,t+\delta^{d,v}(t)\right]}\big] \bigg]. \nonumber
\end{align}
\end{lemma}
Now let us recall \cref{thm:expectation_identity} and prove it, assuming that \cref{lem:k_stops_lemma} holds. Then we will prove \cref{lem:k_stops_lemma}, which completes the proofs of our claims.
\expectationid*
\begin{proof}[Proof of \cref{thm:expectation_identity}]
Examine \cref{eq:k_stops_lemma} when $d\rightarrow \infty$, because we assume the number of events is finite it holds that
\begin{align*}
\lim_{d\rightarrow \infty} \tPr \left( \delta^{d, v}(t) < \delta^{a_{\obs}}(t) \wedge \delta^{a}(t) \wedge T-t \vert \gH_t \right) = 0.
\end{align*}
That is because otherwise, the next treatments should occur after an infinite number of events, and we need to have infinitely many observations before time $T-t$. By convention, we define $\delta^{k,v}(t)=\infty$ whenever there fewer than $k$ events of type $v$ in the interval $(t,T]$. Therefore, also due to the finite amount of events in $(t,T]$, it holds that
\begin{align*}
\lim_{d\rightarrow \infty} \tPr \left( T-t < \delta^{d, v}(t) \wedge \delta^{a_{\obs}}(t) \wedge \delta^{a}(t) \vert \gH_t \right) = \lim_{d\rightarrow \infty} \tPr \left( T-t < \delta^{a_{\obs}}(t) \wedge \delta^{a}(t) \vert \gH_t \right).
\end{align*}
Then we have,
\begin{align*}
\lim_{d\rightarrow \infty}&\E_{\tgH\sim \tPr(\cdot \vert \gH_t)}\bigg[ \\ 
&\sum_{k=1}^{d} \Big( \mathbf{1}\big[\delta^{k-1, v}(t) \leq \delta^a(t) < \delta^{k,v}(t) \wedge \delta^{a_{\obs}}(t) \wedge T-t \big] \E_{P}\big[Y \vert \gH_{t} \cup \tgH^{\setminus a_{\obs}}_{\left(t,t+\delta^{a}(t)\right]}\big] \\
&+\mathbf{1}\big[\delta^{k-1, v}(t) \leq \delta^{a_{\obs}}(t) < \delta^{k, v}(t) \wedge \delta^{a}(t) \wedge T-t \big] \E_{P}\big[Y \vert \gH_{t} \cup \tgH^{\setminus a_{\obs}}_{\left(t,t+\delta^{a_{\obs}}(t)\right]}\big] \Big) \\
&+\mathbf{1}\big[ T-t < \delta^{d,v}(t)\wedge \delta^{a_\obs}(t)\wedge \delta^{a}(t) \big] \E_{P}[Y \vert \gH_{t} \cup \tgH^{\setminus a_{\obs}}_{\left(t,T\right]}] \\
&+\mathbf{1}\big[\delta^{d, v}(t) < \delta^{a_{\obs}}(t) \wedge \delta^{a}(t) \wedge T-t \big] \E_{P}\big[Y \vert \gH_{t} \cup \tgH^{\setminus a_{\obs}}_{\left(t,t+\delta^{d,v}(t)\right]}\big] \bigg] = \\
\lim_{d\rightarrow \infty}&\E_{\tgH\sim \tPr(\cdot \vert \gH_t)}\bigg[ \\ 
&\sum_{k=1}^{d} \Big( \mathbf{1}\big[\delta^{k-1, v}(t) \leq \delta^a(t) < \delta^{k,v}(t) \wedge \delta^{a_{\obs}}(t) \wedge T-t \big] \E_{P}\big[Y \vert \gH_{t} \cup \tgH^{\setminus a_{\obs}}_{\left(t,t+\delta^{a}(t)\right]}\big] \\
&+\mathbf{1}\big[\delta^{k-1, v}(t) \leq \delta^{a_{\obs}}(t) < \delta^{k, v}(t) \wedge \delta^{a}(t) \wedge T-t \big] \E_{P}\big[Y \vert \gH_{t} \cup \tgH^{\setminus a_{\obs}}_{\left(t,t+\delta^{a_{\obs}}(t)\right]}\big] \Big) \\
&+\mathbf{1}\big[ T-t < \delta^{a_\obs}(t)\wedge \delta^{a}(t) \big] \E_{P}[Y \vert \gH_{t} \cup \tgH^{\setminus a_{\obs}}_{\left(t,T\right]}] \bigg]
\end{align*}
Note that the limit when $d\rightarrow \infty$ exists since from \cref{lem:k_stops_lemma} the expectation has the same value for any value of $d$. Next we make two observations:
\begin{itemize}
    \item The conditioning in the $\E_P[Y \vert \ldots]$ terms can be rewritten as $\gH_{t} \cup \tgH^{\setminus a_{\obs}}_{\left(t,t+\delta_{\tgH}(t)\right]}$ for all items. This is because by definition of $\delta_{\tgH}(t)$ as $\min\{u-t : u > t, (u,\cdot)\in{\tgH^{a, a_{\obs}}}\}$, or $\delta_{\tgH}(t)=T-t$ when the set is empty,
    \begin{align*}
        \mathbf{1}\big[\delta^{k-1, v}(t) \leq \delta^a(t) < \delta^{k,v}(t) \wedge \delta^{a_{\obs}}(t) \wedge T-t \big] &= 1 \Rightarrow \delta_{\tgH}(t) = \delta^a(t), \\
        \mathbf{1}\big[\delta^{k-1, v}(t) \leq \delta^{a_{\obs}}(t) < \delta^{k, v}(t) \wedge \delta^{a}(t) \wedge T-t \big] &= 1 \Rightarrow \delta_{\tgH}(t) = \delta^{a_{\obs}}(t), \\
        \mathbf{1}\big[ T-t < \delta^{a_\obs}(t)\wedge \delta^{a}(t) \big] &= 1 \Rightarrow \delta_{\tgH}(t) = T-t.
    \end{align*}
    In the next step we will replace these times with $\delta_{\tgH}(t)$ for all items.
    \item All the events in the indicators are mutually exclusive, and exactly one of them occurs for each $\tgH$. This is since $\delta^{a}(t)\wedge \delta^{a_{\obs}(t)}$ either occurs between the $k-1$ and $k$-th event of type $v$ for some value of $k$, or $\delta^{a}(t)\wedge \delta^{a_{\obs}(t)} > T-t$.
\end{itemize}
Combining these observations into the expressions we
developed for $\E_{P}[Y \vert \gH_t]$, we conclude the proof,
\begin{align*}
\E_P[Y \vert \gH_t]& = \\
\lim_{d\rightarrow \infty} &\E_{\tgH\sim \tPr(\cdot \vert \gH_t)}\bigg[ \\ 
&\sum_{k=1}^{d} \Big( \mathbf{1}\big[\delta^{k-1, v}(t) \leq \delta^a(t) < \delta^{k,v}(t) \wedge \delta^{a_{\obs}}(t) \wedge T-t \big] \E_{P}\big[Y \vert \gH_{t} \cup \tgH^{\setminus a_{\obs}}_{\left(t,t+\delta^{a}(t)\right]}\big] \\
&+\mathbf{1}\big[\delta^{k-1, v}(t) \leq \delta^{a_{\obs}}(t) < \delta^{k, v}(t) \wedge \delta^{a}(t) \wedge T-t \big] \E_{P}\big[Y \vert \gH_{t} \cup \tgH^{\setminus a_{\obs}}_{\left(t,t+\delta^{a_{\obs}}(t)\right]}\big] \Big) \\
&+\mathbf{1}\big[ T-t < \delta^{a_\obs}(t)\wedge \delta^{a}(t) \big] \E_{P}[Y \vert \gH_{t} \cup \tgH^{\setminus a_{\obs}}_{\left(t,T\right]}] \bigg] = \\ 
\lim_{d\rightarrow \infty}&\E_{\tgH\sim \tPr(\cdot \vert \gH_t)}\bigg[ \\ 
&\sum_{k=1}^{d} \Big( \mathbf{1}\big[\delta^{k-1, v}(t) \leq \delta^a(t) < \delta^{k,v}(t) \wedge \delta^{a_{\obs}}(t) \wedge T-t \big] \E_{P}\big[Y \vert \gH_{t} \cup \tgH^{\setminus a_{\obs}}_{\left(t,t+\delta_{\tgH}(t)\right]}\big] \\
&+\mathbf{1}\big[\delta^{k-1, v}(t) \leq \delta^{a_{\obs}}(t) < \delta^{k, v}(t) \wedge \delta^{a}(t) \wedge T-t \big] \E_{P}\big[Y \vert \gH_{t} \cup \tgH^{\setminus a_{\obs}}_{\left(t,t+\delta_{\tgH}(t)\right]}\big] \Big) \\
&+\mathbf{1}\big[ T-t < \delta^{a_\obs}(t)\wedge \delta^{a}(t) \big] \E_{P}[Y \vert \gH_{t} \cup \tgH^{\setminus a_{\obs}}_{\left(t,t+\delta_{\tgH}(t)\right]}] \bigg] = \\
\E_{\tgH\sim \tPr(\cdot \vert \gH_t)}\big[ &\E_{P}[Y \vert \gH_t \cup \tgH^{\setminus a_{\obs}}_{\left(t, t+\delta_{\tgH}(t)\right]}] \big]
\end{align*}
\end{proof}
Next, let us complete the proof of the required lemma that we assumed to hold.
\begin{proof} [Proof of \cref{lem:k_stops_lemma}]
For $d=1$, \cref{eq:k_stops_lemma} is exactly \cref{eq:alg_expect_general} which we already proved in \cref{lem:one_stop_lemma}, and we will proceed by induction. Assume for some $d-1>1$ that \cref{eq:k_stops_lemma} holds, this hypothesis is written below, 
\begin{align} \label{eq:process_proof_induction_hypo}
\E_P[Y \vert \gH_t] &= \E_{\tgH\sim \tPr(\cdot \vert \gH_t)}\Big[ \\
&\sum_{k=1}^{d-1} \Big( \mathbf{1}\big[\delta^{k-1,v}(t) \leq \delta^a(t) < \delta^{k,v}(t) \wedge \delta^{a_{\obs}}(t) \wedge T-t \big] \E_{P}\big[Y \vert \gH_t \cup \tgH^{\setminus a_{\obs}}_{\left( t, t+\delta^a(t)\right]}\big] 
\nonumber \\
&+\mathbf{1}\big[\delta^{k-1, v}(t) \leq \delta^{a_{\obs}}(t) < \delta^{k, v}(t) \wedge \delta^{a}(t) \wedge T-t \big] \E_{P}\big[Y \vert \gH_t \cup \tgH^{\setminus a_{\obs}}_{\left(t, t+\delta^{a_{\obs}}(t)\right]}\big] \nonumber \\
&+\mathbf{1}\big[ T-t < \delta^{d-1,v}(t)\wedge \delta^{a_{\obs}}(t)\wedge \delta^{a}(t) \big] \E_{P}[Y \vert \gH_t \cup \tgH^{\setminus a_{\obs}}_{\left(t, T\right]}]\Big)\nonumber \\
&+\mathbf{1}\big[\delta^{d-1, v}(t) \leq \delta^{a_{\obs}}(t) \wedge \delta^{a}(t) \wedge T-t \big] \E_{P}\big[Y \vert \gH_t \cup \tgH^{\setminus a_{\obs}}_{\left(t, t+ \delta^{d-1,v}(t)\right]}\big] 
\Big]. \nonumber
\end{align}
Using the notation $t_{d-1}=t+\delta^{d-1, v}(t)$ for shorthand, we may rewrite the last summand as
\begin{align*}
    &\E_{\tgH\sim \tPr(\cdot \vert \gH_t)}\bigg[\mathbf{1}\big[\delta^{d-1,v}(t) \leq \delta^{a_{\obs}}(t) \wedge \delta^{a}(t) \wedge T-t \big] \E_{P}\big[Y \vert \gH_{t} \cup \tgH^{\setminus a_{\obs}}_{\left( t, t_{d-1}\right]}\big] 
\bigg] = \\
&\E_{\tgH_{t_{d-1}}\sim \tPr(\cdot \vert \gH_t)}\bigg[ \\
&\E_{\tgH_{(t_{d-1},T]}\sim \tPr(\cdot \vert \tgH_{t_{d-1}})}\bigg[\mathbf{1}\big[\delta^{d-1,v}(t) \leq \delta^{a_{\obs}}(t) \wedge \delta^{a}(t) \wedge T-t\big] \E_{P}\big[Y \vert \gH_t \cup \tgH^{\setminus a_{\obs}}_{\left( t, t_{d-1}\right]}\big] 
\bigg]\bigg] = \\
&\E_{\tgH_{t_{d-1}}\sim \tPr(\cdot \vert \gH_t)}\bigg[\mathbf{1}\big[\delta^{d-1,v}(t) \leq \delta^{a_{\obs}}(t) \wedge \delta^{a}(t) \wedge T-t \big] \E_{P}\big[Y \vert \gH_t \cup \tgH^{\setminus a_{\obs}}_{\left( t, t_{d-1}\right]}\big] 
\bigg].
\end{align*}
The first equality above used the law of total probability, while the second holds since the event 
$\mathbf{1}\big[\delta^{d-1,v}(t) \leq \delta^{a_{\obs}}(t) \wedge \delta^{a}(t) \wedge T-t \big]$
and expectation $\E_{P}\big[Y \vert \gH_t \cup \tgH^{\setminus a_{\obs}}_{\left( t, t_{d-1}\right]}\big]$ only depend on events up to time $t_{d-1}$.
Next we expand the latter term, $\E_{P}\big[Y \vert \gH_t \cup \tgH^{\setminus a_{\obs}}_{\left( t, t_{d-1}\right]}\big]$ according to \cref{lem:one_stop_lemma} and plug-in to the equation above.
\begin{align*}
&\E_{\tgH_{t_{d-1}}\sim \tPr(\cdot \vert \gH_t)}\bigg[\mathbf{1}\big[\delta^{d-1,v}(t) \leq \delta^{a_{\obs}}(t) \wedge \delta^{a}(t) \wedge T-t \big] \E_{P}\big[Y \vert \gH_t \cup \tgH^{\setminus a_{\obs}}_{\left( t, t_{d-1}\right]}\big] 
\bigg] = \\
&\E_{\tgH_{t_{d-1}}\sim \tPr(\cdot \vert \gH_t)}\Bigg[\mathbf{1}\big[\delta^{d-1,v}_{\tgH}(t) \leq \delta_{\tgH}^{a_{\obs}}(t) \wedge \delta_{\tgH}^{a}(t) \wedge T-t \big]\cdot \\
&\quad \E_{\dbtilde{\gH}\sim \tPr\left(\cdot \Big\vert \gH_t \cup \tgH^{\setminus a_{\obs}}_{\left(t, t_{d-1}\right]}\right)}\bigg[
\Big(\mathbf{1}\big[\delta_{\dbtilde{\gH}}^a(t_{d-1}) < \delta_{\dbtilde{\gH}}^v(t_{d-1}) \wedge \delta_{\dbtilde{\gH}}^{a_{\obs}}(t_{d-1})\big]\cdot \\
&\qquad\qquad\qquad\qquad\qquad\E_P\left[ Y \vert \gH_t \cup \tgH^{\setminus a_{\obs}}_{\left( t, t_{d-1}\right]} \cup (t_{d-1}+\delta_{\dbtilde{\gH}}^a(t_{d-1}), a) \right] \Big) + \\
& \qquad\qquad\qquad\qquad\qquad \Big( \mathbf{1}\big[\delta_{\dbtilde{\gH}}^{a_{\obs}}(t_{d-1}) < \delta_{\dbtilde{\gH}}^a(t_{d-1}) \wedge \delta_{\dbtilde{\gH}}^{v}(t_{d-1})\big]\cdot \\
&\qquad\qquad\qquad\qquad\qquad\E_P\left[ Y \vert \gH_{t+\delta_{\dbtilde{\gH}}^{a_{\obs}}(t_{d-1})}=\gH_t \cup \tgH^{\setminus a_{\obs}}_{\left( t, t_{d-1}\right]} \right]\Big) + \\
& \qquad\qquad\qquad\qquad\qquad \Big( \mathbf{1}\big[\delta_{\dbtilde{\gH}}^{a_{\obs}}(t_{d-1}) \wedge \delta_{\dbtilde{\gH}}^a(t_{d-1}) \wedge \delta_{\dbtilde{\gH}}^{v}(t_{d-1}) > T-t_{d-1}\big]\cdot \\
&\qquad\qquad\qquad\qquad\qquad\E_P\left[ Y \vert \gH_{T}=\gH_t \cup \tgH^{\setminus a_{\obs}}_{\left( t, t_{d-1}\right]} \right]\Big) + \\
& \qquad\qquad\qquad\qquad\qquad \Big( \mathbf{1}\big[\delta_{\dbtilde{\gH}}^v(t_{d-1}) < \delta_{\dbtilde{\gH}}^a(t_{d-1}) \wedge \delta_{\dbtilde{\gH}}^{a_{\obs}}(t_{d-1})\big]\cdot \\
&\qquad\qquad\qquad\qquad\qquad\E_P\left[ Y \vert \gH_t \cup \tgH^{\setminus a_{\obs}}_{\left( t, t_{d-1}\right]} \cup (t_{d-1}+\delta_{\dbtilde{\gH}}^v(t_{d-1}), v) \right]\Big)
\bigg]
\Bigg]
\end{align*}
Next we pull out the expectation over $\dbtilde{\gH}$, and then use the law of total probability to turn this into a single expectation over a trajectory $\tgH$ drawn from $\tPr(\cdot \vert \gH_t)$. Each occurrence of $\dbtilde{\gH}$ will then be changed to $\tgH$ accordingly.
\begin{align} \label{eq:lem2_transition1}
&\E_{\tgH_{t_{d-1}}\sim \tPr(\cdot \vert \gH_t)}\bigg[\mathbf{1}\big[\delta^{d-1,v}(t) \leq \delta^{a_{\obs}}(t) \wedge \delta^{a}(t) \wedge T-t \big] \E_{P}\big[Y \vert \gH_t \cup \tgH^{\setminus a_{\obs}}_{\left( t, t_{d-1}\right]}\big] 
\bigg] = \nonumber \\
&\E_{\tgH_{t_{d-1}}\sim \tPr(\cdot \vert \gH_t)}\E_{\dbtilde{\gH}\sim \tPr\left(\cdot \Big\vert \gH_t \cup \tgH^{\setminus a_{\obs}}_{\left(t, t_{d-1}\right]}\right)}\Bigg[\mathbf{1}\big[\delta_{\tgH}^{d-1,v}(t) \leq \delta_{\tgH}^{a_{\obs}}(t) \wedge \delta_{\tgH}^{a}(t) \wedge T -t\big]\cdot \nonumber \\
&\quad\qquad\qquad\qquad\qquad \bigg[
\Big(\mathbf{1}\big[\delta_{\dbtilde{\gH}}^a(t_{d-1}) < \delta_{\dbtilde{\gH}}^v(t_{d-1}) \wedge \delta_{\dbtilde{\gH}}^{a_{\obs}}(t_{d-1})\big]\cdot \nonumber \\
&\qquad\qquad\qquad\qquad\qquad\E_P\left[ Y \vert \gH_t \cup \tgH^{\setminus a_{\obs}}_{\left( t, t_{d-1}\right]} \cup (t_{d-1}+\delta_{\dbtilde{\gH}}^a(t_{d-1}), a) \right] \Big) + \nonumber \\
& \qquad\qquad\qquad\qquad\qquad \Big( \mathbf{1}\big[\delta_{\dbtilde{\gH}}^{a_{\obs}}(t_{d-1}) < \delta_{\dbtilde{\gH}}^a(t_{d-1}) \wedge \delta_{\dbtilde{\gH}}^{v}(t_{d-1})\big]\cdot \nonumber \\
&\qquad\qquad\qquad\qquad\qquad\E_P\left[ Y \vert \gH_{t+\delta_{\dbtilde{\gH}}^{a_{\obs}}(t_{d-1})}=\gH_t \cup \tgH^{\setminus a_{\obs}}_{\left( t, t_{d-1}\right]} \right]\Big) + \nonumber \\
& \qquad\qquad\qquad\qquad\qquad \Big( \mathbf{1}\big[\delta_{\dbtilde{\gH}}^{a_{\obs}}(t_{d-1}) \wedge \delta_{\dbtilde{\gH}}^a(t_{d-1}) \wedge \delta_{\dbtilde{\gH}}^{v}(t_{d-1}) > T-t_{d-1}\big]\cdot \nonumber \\
&\qquad\qquad\qquad\qquad\qquad\E_P\left[ Y \vert \gH_{T}=\gH_t \cup \tgH^{\setminus a_{\obs}}_{\left( t, t_{d-1}\right]} \right]\Big) + \nonumber \\
& \qquad\qquad\qquad\qquad\qquad \Big( \mathbf{1}\big[\delta_{\dbtilde{\gH}}^v(t_{d-1}) < \delta_{\dbtilde{\gH}}^a(t_{d-1}) \wedge \delta_{\dbtilde{\gH}}^{a_{\obs}}(t_{d-1})\big]\cdot \nonumber \\
&\qquad\qquad\qquad\qquad\qquad\E_P\left[ Y \vert \gH_t \cup \tgH^{\setminus a_{\obs}}_{\left( t, t_{d-1}\right]} \cup (t_{d-1}+\delta_{\dbtilde{\gH}}^v(t_{d-1}), v) \right]\Big) \nonumber
\bigg]
\Bigg] = \nonumber \\
&\E_{\tgH\sim \tPr(\cdot \vert \gH_t)}\Bigg[\mathbf{1}\big[\delta_{\tgH}^{d-1,v}(t) \leq \delta_{\tgH}^{a_{\obs}}(t) \wedge \delta_{\tgH}^{a}(t) \wedge T-t \big]\cdot \nonumber \\
&\qquad\qquad\qquad \bigg[
\Big(\mathbf{1}\big[\delta_{\tgH}^a(t_{d-1}) < \delta_{\tgH}^v(t_{d-1}) \wedge \delta_{\tgH}^{a_{\obs}}(t_{d-1})\big]\cdot \nonumber \\
&\qquad\qquad\qquad\qquad\E_P\left[ Y \vert \gH_t \cup \tgH^{\setminus a_{\obs}}_{\left( t, t_{d-1}\right]} \cup (t_{d-1}+\delta_{\tgH}^a(t_{d-1}), a) \right] \Big) + \nonumber \\
&\qquad\qquad\qquad\qquad \Big( \mathbf{1}\big[\delta_{\tgH}^{a_{\obs}}(t_{d-1}) < \delta_{\tgH}^a(t_{d-1}) \wedge \delta_{\tgH}^{v}(t_{d-1})\big]\cdot \nonumber \\
&\qquad\qquad\qquad\qquad\E_P\left[ Y \vert \gH_{t+\delta_{\tgH}^{a_{\obs}}(t_{d-1})}=\gH_t \cup \tgH^{\setminus a_{\obs}}_{\left( t, t_{d-1}\right]} \right]\Big) +  \nonumber \\
&\qquad\qquad\qquad\qquad \Big( \mathbf{1}\big[\delta_{\tgH}^{a_{\obs}}(t_{d-1}) \wedge \delta_{\tgH}^a(t_{d-1}) \wedge \delta_{\tgH}^{v}(t_{d-1}) > T-t_{d-1}\big]\cdot \nonumber \\
&\qquad\qquad\qquad\qquad\E_P\left[ Y \vert \gH_{T}=\gH_t \cup \tgH^{\setminus a_{\obs}}_{\left( t, t_{d-1}\right]} \right]\Big) + \nonumber \\
&\qquad\qquad\qquad\qquad \Big( \mathbf{1}\big[\delta_{\tgH}^v(t_{d-1}) < \delta_{\tgH}^a(t_{d-1}) \wedge \delta_{\tgH}^{a_{\obs}}(t_{d-1})\big]\cdot \nonumber \\
&\qquad\qquad\qquad\qquad\E_P\left[ Y \vert \gH_t \cup \tgH^{\setminus a_{\obs}}_{\left( t, t_{d-1}\right]} \cup (t_{d-1}+\delta_{\tgH}^v(t_{d-1}), v) \right]\Big)
\bigg]
\Bigg]
\end{align}
Let us simplify the multiples of all the indicators that appear in \cref{eq:lem2_transition1}, while dropping the $\tgH$ subscripts since they are clear from context.
\begin{align}\label{eq:lem2_ind_mult}
&\mathbf{1}\big[\delta_{\tgH}^{d-1,v}(t) \leq \delta_{\tgH}^{a_{\obs}}(t) \wedge \delta_{\tgH}^{a}(t) \wedge T-t \big]\cdot \mathbf{1}\big[\delta_{\tgH}^{a}(t_{d-1}) < \delta_{\tgH}^{a_{\obs}}(t_{d-1}) \wedge \delta_{\tgH}^{v}(t_{d-1}) \big] = \nonumber \\
& \quad \mathbf{1}\big[ \delta^{d-1,v}(t) \leq \delta^{a}(t) < \delta^{d,v}(t) \wedge \delta^{a_{\obs}}(t) \wedge T-t \big], \nonumber \\
&\mathbf{1}\big[\delta_{\tgH}^{d-1,v}(t) \leq \delta_{\tgH}^{a_{\obs}}(t) \wedge \delta_{\tgH}^{a}(t) \wedge T-t \big]\cdot \mathbf{1}\big[\delta_{\tgH}^{a_{\obs}}(t_{d-1}) < \delta_{\tgH}^a(t_{d-1}) \wedge \delta_{\tgH}^{v}(t_{d-1}) \big] = \nonumber \\
& \quad \mathbf{1}\big[ \delta^{d-1,v}(t) \leq \delta^{a_{\obs}}(t) < \delta^{d,v}(t) \wedge \delta^{a}(t) \wedge T-t \big], \nonumber \\
&\mathbf{1}\big[\delta_{\tgH}^{d-1,v}(t) \leq \delta_{\tgH}^{a_{\obs}}(t) \wedge \delta_{\tgH}^{a}(t) \wedge T-t \big]\cdot \mathbf{1}\big[\delta_{\tgH}^{a_{\obs}}(t_{d-1}) \wedge \delta_{\tgH}^a(t_{d-1}) \wedge \delta_{\tgH}^{v}(t_{d-1}) > T-t_{d-1}\big] = \nonumber \\
& \quad \mathbf{1}\big[ \delta^{d,v}(t) \wedge \delta^{a_{\obs}}(t) \wedge \delta^{a}(t) > T-t \big], \nonumber \\
&\mathbf{1}\big[\delta_{\tgH}^{d-1,v}(t) \leq \delta_{\tgH}^{a_{\obs}}(t) \wedge \delta_{\tgH}^{a}(t) \wedge T-t \big]\cdot \mathbf{1}\big[\delta_{\tgH}^{v}(t_{d-1}) < \delta_{\tgH}^a(t_{d-1}) \wedge \delta_{\tgH}^{a_{\obs}}(t_{d-1}) \big] = \nonumber \\
& \quad \mathbf{1}\big[ \delta^{d,v}(t) < \delta^{a_{\obs}}(t) \wedge \delta^{a}(t) \wedge T-t \big].
\end{align}
Switching these equalities from \ref{eq:lem2_ind_mult} into \cref{eq:lem2_transition1}, we get
\begin{align*}
&\E_{\tgH_{t_{d-1}}\sim \tPr(\cdot \vert \gH_t)}\bigg[\mathbf{1}\big[\delta^{d-1,v}(t) \leq \delta^{a_{\obs}}(t) \wedge \delta^{a}(t) \wedge T-t \big] \E_{P}\big[Y \vert \gH_t \cup \tgH^{\setminus a_{\obs}}_{\left( t, t_{d-1}\right]}\big] 
\bigg] = \nonumber \\
&\E_{\tgH\sim \tPr(\cdot \vert \gH_t)}\Bigg[\Big(\mathbf{1}\big[ \delta^{d-1,v}(t) \leq \delta^{a}(t) < \delta^{d,v}(t) \wedge \delta^{a_{\obs}}(t) \wedge T-t \big]\cdot \\
&\qquad\qquad\quad~ \E_P\left[ Y \vert \gH_t \cup \tgH^{\setminus a_{\obs}}_{\left( t, t_{d-1}\right]} \cup (t_{d-1}+\delta_{\tgH}^a(t_{d-1}), a) \right] \Big) + \nonumber \\
&\qquad\qquad\quad~ \Big( \mathbf{1}\big[ \delta^{d-1,v}(t) \leq \delta^{a_{\obs}}(t) < \delta^{d,v}(t) \wedge \delta^{a}(t) \wedge T-t \big]\cdot \nonumber \\
&\qquad\qquad\quad~ \E_P\left[ Y \vert \gH_{t+\delta_{\tgH}^{a_{\obs}}(t_{d-1})}=\gH_t \cup \tgH^{\setminus a_{\obs}}_{\left( t, t_{d-1}\right]} \right]\Big) + \\
&\qquad\qquad\quad~ \Big( \mathbf{1}\big[ \delta^{d,v}(t) \wedge \delta^{a_{\obs}}(t) \wedge \delta^{a}(t) > T-t \big]\cdot \nonumber \\
&\qquad\qquad\quad~ \E_P\left[ Y \vert \gH_{T}=\gH_t \cup \tgH^{\setminus a_{\obs}}_{\left( t, t_{d-1}\right]} \right]\Big) + \nonumber \\
&\qquad\qquad\quad~\Big(\mathbf{1}\big[ \delta^{d,v}(t) < \delta^{a_{\obs}}(t) \wedge \delta^{a}(t) \wedge T-t \big]\cdot \\
&\qquad\qquad\quad~\E_P\left[ Y \vert \gH_t \cup \tgH^{\setminus a_{\obs}}_{\left( t, t_{d-1}\right]} \cup (t_{d-1}+\delta_{\tgH}^v(t_{d-1}), v) \right]\Big)
\bigg]
\Bigg].
\end{align*}
Then it is easy to deduce that the conditioning sets can be simplified as follows,
\begin{align*}
&\E_{\tgH_{t_{d-1}}\sim \tPr(\cdot \vert \gH_t)}\bigg[\mathbf{1}\big[\delta^{d-1,v}(t) \leq \delta^{a_{\obs}}(t) \wedge \delta^{a}(t) \wedge T-t \big] \E_{P}\big[Y \vert \gH_t \cup \tgH^{\setminus a_{\obs}}_{\left( t, t_{d-1}\right]}\big] 
\bigg] = \nonumber \\
&\E_{\tgH\sim \tPr(\cdot \vert \gH_t)}\Bigg[\Big(\mathbf{1}\big[ \delta^{d-1,v}(t) \leq \delta^{a}(t) < \delta^{d,v}(t) \wedge \delta^{a_{\obs}}(t) \wedge T-t \big]\cdot \\
&\qquad\qquad\quad~ \E_P\left[ Y \vert \gH_t \cup \tgH^{\setminus a_{\obs}}_{\left( t, \delta^{a}(t)\right]} \right] \Big) + \nonumber \\
&\qquad\qquad\quad~ \Big( \mathbf{1}\big[ \delta^{d-1,v}(t) \leq \delta^{a_{\obs}}(t) < \delta^{d,v}(t) \wedge \delta^{a}(t) \wedge T-t \big]\cdot \nonumber \\
&\qquad\qquad\quad~ \E_P\left[ Y \vert \gH_t \cup \tgH^{\setminus a_{\obs}}_{\left( t, t+\delta^{a_{\obs}}(t)\right]} \right]\Big) +  \nonumber \\
&\qquad\qquad\quad~ \Big( \mathbf{1}\big[ \delta^{d,v}(t) \wedge \delta^{a_{\obs}}(t) \wedge \delta^{a}(t) > T-t \big]\cdot \nonumber \\
&\qquad\qquad\quad~ \E_P\left[ Y \vert \gH_{T}=\gH_t \cup \tgH^{\setminus a_{\obs}}_{\left( t, T\right]} \right]\Big) + \nonumber \\
&\qquad\qquad\quad~\Big(\mathbf{1}\big[ \delta^{d,v}(t) < \delta^{a_{\obs}}(t) \wedge \delta^{a}(t) \wedge T-t \big]\cdot \\
&\qquad\qquad\quad~\E_P\left[ Y \vert \gH_t \cup \tgH^{\setminus a_{\obs}}_{\left( t, t+\delta^{d, v}(t)\right]} \right]\Big)
\bigg]
\Bigg].
\end{align*}
Plugging this back into \cref{eq:process_proof_induction_hypo} and using the linearity of expectation gives us exactly the equality in \cref{eq:k_stops_lemma} and concludes the proof.
\end{proof}

\input{sections/discrete_time_temp}

\section{Additional Discussion on Related Work} \label{app:related_work}
As outlined in \cref{sec:related_work}, several techniques have been proposed for scalable estimation of causal effects in sequential decision-making, with more limited development in the case of irregular observation times. One set of approaches \citep{Bica2020Estimating, melnychuk2022causal, lim2018forecasting}, that only apply to discrete time processes and static policies, can be roughly characterized as follows. A prediction model $f(\gH_t, \gH^a_T; \vtheta)$ for the outcome $Y$ is learned, where $\gH_t$ is the observed history of events and $\gH^a_T$ is the set of future treatments we would like to reason about. That is, in our notation we would like $f(\gH_t, \gH^a_T; \vtheta)$ to estimate $\E_{P_{\gH^a_T}}\left[ Y \vert \gH_t \right]$, where $P_{\gH^a_T}$ assigns the treatments in $\gH^a_T$ w.p. $1$. In potential outcomes notation, this corresponds to $\E\left[ Y^{\rva} \vert \gH_t \right]$, where $Y^\rva$ is a random variable that outputs the outcome under a set of static future treatments $\rva$. All methods involve learning a representation of history $\rmZ_t = \phi(\gH_t; \eta)$, and combine two important elements for achieving correct estimates.
\begin{enumerate}
    \item To yield correct causal estimates under an observational distribution that is not sequentially randomized, methods either estimate products of propensity weights \citep{lim2018forecasting}, or add a loss to make $\rmZ_t$ non-predictive of the treatment $A_t$, $\phi$ is then called a balancing representation.
    \item To facilitate prediction of $Y$ under a set of future treatments in the interval $(t,T]$, either $\phi$ is taken as a sequence model, or a separate ``decoder" network is learned \citep{lim2018forecasting, Bica2020Estimating}. A sequence model is trained with inputs where $\gH^{x,y}_{i, T} \setminus \gH^{x,y}_{i, t}$, i.e. the covariates in a projection interval $(t, T]$ are masked, while the decoder takes $\rmZ_t$ and $\gH^a_T$ as inputs. Both are trained to predict the outcome $Y$ and serve as an estimator for $\E_{P_{\obs}}[Y \vert \rmZ_t, \gH_T^a]$, which recovers the correct causal effect under sequential exchangeability. Notice that these techniques preclude estimation with dynamic treatments, i.e. policies. 
\end{enumerate}

For irregular sampling, \citet{seedat2022continuous} follow the same recipe but choose a neural CDE architecture. This interpolates the latent path $\rmZ_t$ in intervals between jump times of the processes, and is shown empirically to be more suitable when working with data that is subsampled from a complete trajectory of features in continuous time. The solution is not equipped to estimate interventions on continuous treatment times (in our notation, $\lambda^a$). As mentioned earlier, \citet{vanderschueren2023accounting} handle informative sampling times with inverse weighting based on the intensity $\lambda$. However, this is a different problem setting from ours, as they do not seek to intervene on sampling times but wish to solve a case where outcomes, features and treatments always jump simultaneously. In our setting, intervening on $\lambda^a$ with such simultaneous jumps would result in $\lambda^{x,y}_{\obs} \neq \lambda^{x,y}$, which is not the focus of our work. Finally, we also note the required assumption for causal validity that is claimed in these works is roughly $P_\obs(A_t=a_t \vert \gH_T) = P_\obs(A_t=a_t \vert \gH_t)$. The assumption is unreasonable since $\gH_T$ includes future \emph{factual} outcomes that depend on the taken action, instead of the more standard exchangeability assumption that posits independence of \emph{potential} outcomes.

The G-estimation solution of \citep{pmlr-v158-li21a} for discrete time decision processes fits models for both $\pi_{\obs}(A(t) \vert \gH_{t-1}, X(t))$, and $P_{\obs}(X(t), Y(t) \vert \gH_{t-1})$. Then at inference time, they replace $\pi_{\obs}$ with the desired policy $\pi$ and estimate trajectories or conditional expectations of $Y$ with monte-carlo simulations. A straightforward generalization of this approach to decision point processes can be devised by fitting the intensities $\lambda_{\obs}$ and replacing $\lambda^a$ for inference.
While we believe that this is an interesting direction for future work, we do not pursue it further in our experiments, since developing architectures and methods for learning generative models under irregular sampling deserves a dedicated and in-depth exploration.


%% file: sections/discrete_time_temp.tex
\subsection{Discrete Time Version}\label{sec:tower_expectation_discrete}
For the discrete-time version we keep a similar notation, but take time increments of $1$ and call the target policy $\pi$, which takes a history of the process and outputs a distribution over possible treatments. The trajectory $\gH$ now simplifies to the form $\{(\rvx_1, \rvy_1, \rva_1), (\rvx_2, \rvy_2, \rva_2), \ldots, (\rvx_T, \rvy_T, \rva_T)\}$ and similarly for the history $\gH_t$. The analogous claim to \cref{thm:expectation_identity} for these decision processes follows from the lemma we prove below by setting $d=T-t$.
\begin{lemma}
For any $\gH$, $t\in{[0, T)}$ and $1 \leq d \leq T-t$ such that $\gH_t$ is measurable w.r.t $P$ we have that
\begin{align} \label{eq:recurs_discrete}
    \E_P[Y \vert \gH_t] = &\E_{\tgH\sim \tPr(\cdot \vert \gH_t)}\Bigg[ \nonumber \\
    & \sum_{k=1}^{d}\Big( \mathbf{1}_{a_{t+k}\neq a_{\obs, t+k}}\prod_{i=1}^{k-1}{\mathbf{1}_{a_{\obs,t+k-i}=a_{t+k-i}}}\E_{P}[Y \vert \gH_{t+k} = \gH_t \cup \tgH^{\setminus a_{\obs}}_{[t+1, t+k]}]\Big) \nonumber \\
+&\prod_{k=1}^{d}\mathbf{1}_{a_{t+k}=a_{\obs, t+k}}\E_P[Y \vert \gH_{t+d} = \gH_t \cup \tgH^{\setminus a_{\obs}}_{[t+1, t+d]}]\Bigg].
\end{align}
Note that for $k=1$, we define $\prod_{i=1}^{k-1}{\mathbf{1}_{a_{t+k-i}=a_{\obs, t+k-i}}}=1$.
\end{lemma}
\begin{proof}
Throughout the proof we will use conditioning on $\tgH_{t+k}$ (where $0 \leq k \leq d$) as a shorthand for conditioning on $\gH_{t+k} = \gH_t \cup \tgH^{\setminus a_{\obs}}_{[t+1, t+k]}$, as the meaning is clear from context.
We will prove the claim by induction on $d$. The base case for $d=1$ follows simply from observing that the value drawn for $a_{\obs, t+1}$ does not change the items inside the expectation $\E_{\tgH\sim\tPr(\cdot \vert \gH_t)}[\ldots]$. Let us write this down in detail.
\begin{align} \label{eq:disc_proof_eq1}
    &\E_{\tgH\sim \tPr( \cdot \vert \gH_t)} \Bigg[ \mathbf{1}_{{a}_{t+1}\neq a_{\obs, t+1}}\E_{P}[Y \vert \tgH_{t+1}] +\mathbf{1}_{{a}_{t+1}=a_{\obs, t+1}}\E_P[Y \vert \tgH_{t+1}]\Bigg] = \nonumber \\
    &\E_{\tgH\sim \tPr( \cdot \vert \gH_t)} \Bigg[ \E_{P}[Y \vert \tgH_{t+1}]\Bigg] = \nonumber \\
    &\E_{\tgH_{t+1}\sim P_\obs( \cdot \vert \gH_t)} \Bigg[ \E_{P}[Y \vert \tgH_{t+1}]\Bigg]= \nonumber \\
    &\E_{\rvx_{t+1}, \rvy_{t+1}\sim P_{\obs}( \cdot \vert \gH_t), a_{t+1}\sim \pi(\cdot \vert \gH_t, \rvx_{t+1}, \rvy_{t+1})} \Bigg[ \nonumber \\
&\quad \E_{a_{\obs, t+1}\sim\pi_{\obs}(\cdot \vert \gH_{t}, \rvx_{t+1}, \rvy_{t+1})}\Bigg[ \E_P[Y \vert \tgH_{t+1}]\Bigg]\Bigg] = \nonumber \\
&\E_{\rvx_{t+1}, \rvy_{t+1}\sim P_{\obs}( \cdot \vert \gH_t), a_{t+1}\sim \pi(\cdot \vert \gH_t, \rvx_{t+1}, \rvy_{t+1})} \Bigg[  \E_P[Y \vert \tgH_{t+1}]\Bigg] = \nonumber \\
&\E_{\rvx_{t+1}, \rvy_{t+1}\sim P( \cdot \vert \gH_t), a_{t+1}\sim \pi(\cdot \vert \gH_t, \rvx_{t+1}, \rvy_{t+1})} \Bigg[  \E_P[Y \vert \tgH_{t+1}]\Bigg] = \nonumber \\
&\E_{P}[Y \vert \gH_t].
\end{align}
The first equality, as we argued earlier is due to the conditioning set $\tgH_{t+1}$ not depending on $a_{\obs, t+1}$ (as we defined earlier, it refers to $\gH_t \cup \tgH^{\setminus a_{\obs}}_{[t+1, t+1]}$). Intuitively, this is true because we only expand the expectation one step forward in time. In the identity we wish to prove, \cref{eq:recurs_discrete}, the $a_{\obs}$ treatments do change the item within $\E_{\tgH\sim\tPr(\cdot \vert \gH_t)}[\ldots]$ since they determine the earliest disagreement time. The second equality is obtained by marginalizing over the sampled trajectories after time $t+1$, as they are also not included in $\tgH_{t+1}$. Then the third equality writes the sampling of $\rvy_{t+1}, \rvx_{t+1}, a_{t+1}, a_{\obs, t+1}$ explicitly, and the fourth marginalizes over $a_{\obs, t+1}$ as we already mentioned it does not appear in the expectation. Then we use the equality $P(X_{t+1}, Y_{t+1} \vert \gH_t) = P_{\obs}(X_{t+1}, Y_{t+1} \vert \gH_t)$, to write the expectation as sampling $\rvx_{t+1}, \rvy_{t+1}, a_{t+1}$ from $P$. Finally, we use the tower property of conditional expectation and arrive at the desired expression.
Next, assume that the claim holds for some $d-1$. We write down the induction hypothesis and then marginalize over all the event after time $t+d-1$ in $\tgH$, as they do not appear in the arguments of the expectation.
\begin{align} \label{eq:disc_proof_eq2}
\E_P[Y \vert \gH_t] = \E_{\tgH\sim \tPr(\cdot \vert \gH_t)}\Bigg[&\sum_{k=1}^{d-1}\left( \mathbf{1}_{a_{t+k}\neq a_{\obs, t+k}}\prod_{i=1}^{k-1}{\mathbf{1}_{a_{t+k-i}=a_{\obs, t+k-i}}}\cdot \E_{P}[Y \vert \tgH_{t+k}]\right) \nonumber \\
+&\prod_{k=1}^{d-1}\mathbf{1}_{a_{t+k}=a_{\obs, t+k}} \cdot \E_P[Y \vert \tgH_{t+d-1}]\Bigg] \nonumber \\
= \E_{\tgH_{t+d-1}\sim \tPr(\cdot \vert \gH_t)}\Bigg[&\sum_{k=1}^{d-1}\left( \mathbf{1}_{a_{t+k}\neq a_{\obs, t+k}}\prod_{i=1}^{k-1}{\mathbf{1}_{a_{t+k-i}=a_{\obs, t+k-i}}}\cdot \E_{P}[Y \vert \tgH_{t+k}]\right) \nonumber \\
+&\prod_{k=1}^{d-1}\mathbf{1}_{a_{t+k}=a_{\obs, t+k}} \cdot \E_P[Y \vert \tgH_{t+d-1}]\Bigg].
\end{align}
As in the base case, we can expand the expectation $\E[Y \vert \tgH_{t+d-1}]$ one time step forward,
\begin{align} \label{eq:total_expectation}
\E_{P}[Y \vert \tgH_{t+d-1}] = \E_{\tgH_{t+d} \sim \tPr(\cdot \vert \tgH_{t+d-1})}\bigg[  &\E_{P}\bigg[Y \vert \tgH_{t+d} \bigg]\cdot \mathbf{1}_{{a}_{t+d}\neq a_{\obs, t+d}} \\
+& \E_{P}\bigg[Y \vert \tgH_{t+d} \bigg]\cdot \mathbf{1}_{{a}_{t+d}=a_{\obs, t+d}}  \bigg]. \nonumber
\end{align}
Now plug this in \cref{eq:disc_proof_eq2} to obtain
\begin{align*}
\E_{P}[Y \vert \gH_t] = \:&\E_{\tgH_{t+d-1}\sim \tPr(\cdot \vert \gH_t)}\Bigg[\sum_{k=1}^{d-1}\left( \mathbf{1}_{a_{t+k}\neq a_{\obs, t+k}}\prod_{i=1}^{k-1}{\mathbf{1}_{a_{t+k-i}=a_{\obs, t+k-i}}}\cdot \E_{P}[Y \vert \tgH_{t+k}]\right) \nonumber \\
+&\prod_{k=1}^{d-1}\mathbf{1}_{a_{t+k}=a_{\obs, t+k}} \cdot \Bigg( \E_{\tgH_{t+d} \sim \tPr(\cdot \vert \tgH_{t+d-1})}\bigg[  \E_{P}\bigg[Y \vert \tgH_{t+d} \bigg]\cdot \mathbf{1}_{{a}_{t+d}\neq a_{\obs, t+d}} \\
+& \E_{P}\bigg[Y \vert \tgH_{t+d} \bigg]\cdot \mathbf{1}_{{a}_{t+d}=a_{\obs, t+d}}  \bigg]\Bigg)\Bigg] \\
=\:& \E_{\tgH_{t+d-1}\sim \tPr(\cdot \vert \gH_t)}\Bigg[\E_{\tgH_{t+d} \sim \tPr(\cdot \vert \tgH_{t+d-1})}\Bigg[ \\
&\sum_{k=1}^{d-1}\left( \mathbf{1}_{a_{t+k}\neq a_{\obs, t+k}}\prod_{i=1}^{k-1}{\mathbf{1}_{a_{t+k-i}=a_{\obs, t+k-i}}}\cdot \E_{P}[Y \vert \tgH_{t+k}]\right) \nonumber \\
+&\prod_{k=1}^{d-1}\mathbf{1}_{a_{t+k}=a_{\obs, t+k}} \cdot \Bigg(   \E_{P}\bigg[Y \vert \tgH_{t+d} \bigg]\cdot \mathbf{1}_{{a}_{t+d}\neq a_{\obs, t+d}} \\
+& \E_{P}\bigg[Y \vert \tgH_{t+d} \bigg]\cdot \mathbf{1}_{{a}_{t+d}=a_{\obs, t+d}}  \Bigg)\Bigg]\Bigg] \\
=\:& \E_{\tgH_{t+d}\sim \tPr(\cdot \vert \gH_t)}\Bigg[ \sum_{k=1}^{d-1}\left( \mathbf{1}_{a_{t+k}\neq a_{\obs, t+k}}\prod_{i=1}^{k-1}{\mathbf{1}_{a_{t+k-i}=a_{\obs, t+k-i}}}\cdot \E_{P}[Y \vert \tgH_{t+k}]\right) \nonumber \\
+&\prod_{k=1}^{d-1}\mathbf{1}_{a_{t+k}=a_{\obs, t+k}} \cdot \Bigg(   \E_{P}\bigg[Y \vert \tgH_{t+d} \bigg]\cdot \mathbf{1}_{{a}_{t+d}\neq a_{\obs, t+d}} \\
+& \E_{P}\bigg[Y \vert \tgH_{t+d} \bigg]\cdot \mathbf{1}_{{a}_{t+d}=a_{\obs, t+d}}  \Bigg)\Bigg] \\
=\:& \E_{\tgH_{t+d}\sim \tPr(\cdot \vert \gH_t)}\Bigg[ \sum_{k=1}^{d}\left( \mathbf{1}_{a_{t+k}\neq a_{\obs, t+k}}\prod_{i=1}^{k-1}{\mathbf{1}_{a_{t+k-i}=a_{\obs, t+k-i}}}\cdot \E_{P}[Y \vert \tgH_{t+k}]\right) \nonumber \\
+& \prod_{k=1}^{d}{\mathbf{1}_{a_{t+k}=a_{\obs, t+k}}} \cdot \E_{P}\bigg[Y \vert \tgH_{t+d} \bigg]  \Bigg)\Bigg] \\
=\:& \E_{\tgH\sim \tPr(\cdot \vert \gH_t)}\Bigg[ \sum_{k=1}^{d}\left( \mathbf{1}_{a_{t+k}\neq a_{\obs, t+k}}\prod_{i=1}^{k-1}{\mathbf{1}_{a_{t+k-i}=a_{\obs, t+k-i}}}\cdot \E_{P}[Y \vert \tgH_{t+k}]\right) \nonumber \\
+& \prod_{k=1}^{d}{\mathbf{1}_{a_{t+k}=a_{\obs, t+k}}} \cdot \E_{P}\bigg[Y \vert \tgH_{t+d} \bigg]  \Bigg)\Bigg]
\end{align*}
The equality between the first and last item is exactly the expression we wish to prove. The first transition plugs \cref{eq:total_expectation} into \cref{eq:disc_proof_eq2}, the second pushes the expectation over $\rvx_{t+d}, \rvy_{t+d}, a_{t+d}, a_{\obs, t+d}$ outside, the third uses the law of total expectation. Then we rearrange
\begin{align*}
&\prod_{k=1}^{d-1}\mathbf{1}_{a_{t+k}=a_{\obs, t+k}} \cdot \Bigg(   \E_{P}\bigg[Y \vert \tgH_{t+d} \bigg]\cdot \mathbf{1}_{{a}_{t+d}\neq a_{\obs, t+d}} 
+ \E_{P}\bigg[Y \vert \tgH_{t+d} \bigg]\cdot \mathbf{1}_{{a}_{t+d}=a_{\obs, t+d}} \Bigg) \\
=& \Bigg( \mathbf{1}_{{a}_{t+d}\neq a_{\obs, t+d}} \cdot \prod_{k=1}^{d-1}\mathbf{1}_{a_{t+k}=a_{\obs, t+k}} \cdot    \E_{P}\bigg[Y \vert \tgH_{t+d} \bigg] \Bigg) 
+ \Bigg(\prod_{k=1}^{d}\mathbf{1}_{a_{t+k}=a_{\obs, t+k}} \cdot\E_{P}\bigg[Y \vert \tgH_{t+d} \bigg] \Bigg),
\end{align*}
and push the first item into the summation over $k$.
\end{proof}
To obtain the identity with the earliest disagreement time, set $d=T-t$ in the statement we proved, and note that for any value of $\tgH$ in the expectation above, exactly one of the following items equals $1$,
\begin{align*}
\mathbf{1}_{{a}_{t+k}\neq a_{\obs, t+k}} \cdot \prod_{i=1}^{k-1}\mathbf{1}_{a_{t+k-i}=a_{\obs, t+k-i}} \text{ for } k\in{[T-t]}, \text{ and } \prod_{k=1}^{d}\mathbf{1}_{a_{t+k}=a_{\obs, t+k}},
\end{align*}
and the value of $k$ for which the above item equals $1$ is $k=\delta_{\tgH}(t)$ (and $T$ when $\prod_{k=1}^{d}\mathbf{1}_{a_{t+k}=a_{\obs, t+k}}=1$), hence $t+\delta_{\tgH}(t)$ is the earliest disagreement time, and \cref{eq:recurs_discrete} reduces to \cref{eq:expectation_identity} as claimed.